\documentclass{article}

\usepackage{arxiv}

\usepackage[utf8]{inputenc} 
\usepackage[T1]{fontenc}    
\usepackage{hyperref}       
\usepackage{url}            
\usepackage{booktabs}       
\usepackage{amsfonts}       
\usepackage{nicefrac}       
\usepackage{microtype}      
\usepackage{amssymb, amsmath}
\usepackage{cleveref}       
\usepackage{lipsum}         
\usepackage{graphicx}
\usepackage{doi}

\title{MaGRITTe: Manipulative and Generative 3D Realization from Image, Topview and Text}

\author{Takayuki Hara\\
	The University of Tokyo\\
	\texttt{hara@mi.t.u-tokyo.ac.jp} \\
	\And
        Tatsuya Harada\\
	The University of Tokyo / RIKEN\\
	\texttt{harada@mi.t.u-tokyo.ac.jp} \\
}

\renewcommand{\shorttitle}{MaGRITTe: Manipulative and Generative 3D Realization from Image, Topview and Text}

\hypersetup{
pdftitle={MaGRITTe: Manipulative and Generative 3D Realization from Image, Topview and Text},
pdfauthor={Takayuki Hara, Tatsuya Harada},
}

\newcommand{\putFigWpdf}[3]{%
  \begin{figure}[tb]%
    \centering%
    \includegraphics[width=#3]{fig/#1.pdf}%
    \caption{#2}%
    \label{fig:#1}%
  \end{figure}}

\newcommand{\putFigWWpdf}[3]{%
  \begin{figure*}[h]%
    \centering%
    \includegraphics[width=#3]{fig/#1.pdf}%
    \caption{#2}%
    \label{fig:#1}%
  \end{figure*}}

\begin{document}
\maketitle

\begin{abstract}

The generation of 3D scenes from user-specified conditions offers a promising avenue for alleviating the production burden in 3D applications. Previous studies required significant effort to realize the desired scene, owing to limited control conditions. We propose a method for controlling and generating 3D scenes under multimodal conditions using partial images, layout information represented in the top view, and text prompts. Combining these conditions to generate a 3D scene involves the following significant difficulties: (1) the creation of large datasets, (2) reflection on the interaction of multimodal conditions, and (3) domain dependence of the layout conditions. We decompose the process of 3D scene generation into 2D image generation from the given conditions and 3D scene generation from 2D images. 2D image generation is achieved by fine-tuning a pretrained text-to-image model with a small artificial dataset of partial images and layouts, and 3D scene generation is achieved by layout-conditioned depth estimation and neural radiance fields (NeRF), thereby avoiding the creation of large datasets. The use of a common representation of spatial information using 360-degree images allows for the consideration of multimodal condition interactions and reduces the domain dependence of the layout control. The experimental results qualitatively and quantitatively demonstrated that the proposed method can generate 3D scenes in diverse domains, from indoor to outdoor, according to multimodal conditions. A project website with supplementary video is here \href{https://hara012.github.io/MaGRITTe-project}{https://hara012.github.io/MaGRITTe-project}.
\end{abstract}

\putFigWWpdf{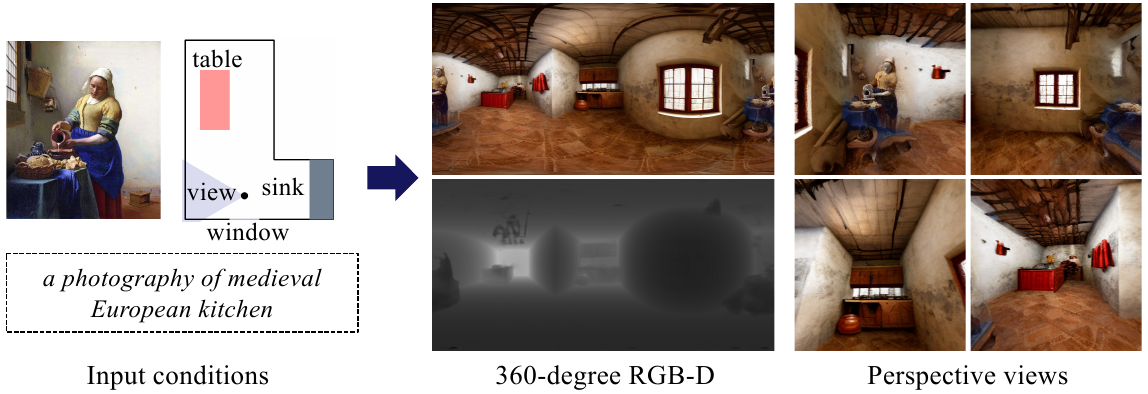}{From a given partial image, layout information represented in top view, and text prompts, our method generates a 3D scene represented by the 360-degree RGB-D, and NeRF. Free perspective views can be rendered from the NeRF model.}{150mm}

\section{Introduction}
\label{sec:intro}

3D scene generation under user-specified conditions is a fundamental task in the fields of computer vision and graphics. In particular, the generation of 3D scenes extending in all directions from the observer's viewpoint is a promising technology that reduces the burden and time of creators and provides them with new ideas for creation in 3D applications such as VR/AR, digital twins, and the metaverse.

In recent years, 3D scene generation under user-specified conditions using generative models  \cite{VAE,VAE2,GAN,transformer,diff_model1,diff_model2} has been extensively studied. A wide range of methods exist for generating 3D scenes from parital images \cite{devries2021unconstrained,gaudi,du2023cross,cheng2023sparsegnv}, layout information such as floor plans and bird's-eye views \cite{Vidanapathirana2021Plan2Scene,bahmani2023cc3d,kim2023nfldm,yang2023bevcontrol,chen2023sd,schult24controlroom3d}, and text prompts \cite{wang2023prolificdreamer,shi2023mvdream,hoellein2023text2room,stan2023ldm3d}.  However, these methods are limited by the conditions they can take as input, making it difficult to generate the 3D scene intended by the user. This is due to the fact that each condition has its own advantages and disadvantages. For example, when partial images are given, it is possible to present a detailed appearance; however, it is difficult to create information outside the image; when a layout is given, it is possible to accurately describe object alignment but not to specify a detailed appearance; when text is given as a condition, it is suitable for specifying the overall context; however, it is difficult to determine the exact shape and appearance of objects.

Considering these problems, we propose a method for generating 3D scenes by simultaneously providing a combination of three conditions: partial images, layout information represented in the top view, and text prompts  (\cref{fig:summary_360rgbd_gen}). This approach aims to compensate for the shortcomings of each condition in a complementary manner, making it easier to create the 3D scenes intended by the creator. That is, details of appearance from partial images, shape and object placement from layout information, and overall context can be controlled using text prompts.

Integrating partial images, layouts, and texts to control a 3D scene involves the following significant difficulties that cannot be addressed by a simple combination of existing methods: (1) creation of large datasets, (2) reflection of the interaction of multimodal conditions, and (3) domain dependence of the layout representations. To overcome these difficulties, we initially decomposed the process of 3D scene generation into two steps: 2D image generation from the given conditions and 3D generation from 2D images. For 2D image generation, our approach is to create small artificial datasets for partial images and layout conditions and fine-tune the text-to-image model trained on a large dataset. We then generated a 3D scene from a 2D image using layout-conditioned monocular depth estimation and training NeRF \cite{nerf}. This approach eliminates the need to create large datasets of 3D scenes. This study aimed to improve scene consistency and reduce computational costs using 360-degree images for 2D image generation. To address the second issue, which reflects the interaction of multimodal conditions, we encoded the input conditions into a common latent space in the form of equirectangular projection (ERP) for 360-degree images. To address the third issue of domain dependence of layout representations, we present a framework for incorporating domain-specific top-view representations with less effort by converting them into more generic intermediate representations of depth and semantic maps in ERP format. This allows for generating various scenes from indoor to outdoor by simply replacing the converter.

The contributions of this study are as follows:
 \begin{itemize}
\item We introduce a method to control and generate 3D scenes from partial images, layouts, and texts, complementing the advantages of each condition.
\item We present a method that avoids the need for creating large datasets by fine-tuning a pre-trained large-scale text-to-image model with a small artificial dataset of partial images and layouts for 2D image generation, and by generating 3D scenes from 2D images through layout-conditioned depth estimation and training NeRF. 
\item We address the integration of different modalities by converting the input information into ERP format, passing it through an encoder, and embedding the information in the same latent space. 
\item We present a framework for generating various scenes from indoor to outdoor with a module for converting top view layout representations into depth maps and semantic maps in ERP format.
\item Experimental results validate that the proposed method can generate 3D scenes with controlled appearance, geometry, and overall context based on input information, even beyond the dataset used for fine-tuning.
\end{itemize}

\section{Related Work}

\subsection{3D Scene Generation}

3D scene generation involves the creation of a model of a 3D space that includes objects and backgrounds, based on user-specified conditions. In recent years, the use of generative models, such as VAE \cite{VAE,VAE2}, GAN \cite{GAN}, autoregressive models \cite{transformer}, and diffusion models \cite{diff_model1,diff_model2}, has made rapid progress. There are methods to generate a 3D scene from random variables \cite{lin2023infinicity,Chai2023persistent}, from one or a few images \cite{devries2021unconstrained,gaudi,rgbd2,du2023cross,cheng2023sparsegnv}, from layout information such as floor plans \cite{Vidanapathirana2021Plan2Scene,bahmani2023cc3d}, bird's-eye views (semantic maps in top view) \cite{kim2023nfldm,yang2023bevcontrol}, terrain maps \cite{chen2023sd} and 3D proxies \cite{schult24controlroom3d}, and as well as from text prompts \cite{wang2023prolificdreamer,shi2023mvdream,hoellein2023text2room,stan2023ldm3d,SceneScape}. However, each method has its own advantages and disadvantages in terms of scene control characteristics, and it is difficult to generate a 3D scene that appropriately reflects the intentions. We propose a method to address these challenges by integrating partial images, layout information, and text prompts as input conditions in a complementary manner. Furthermore, while layout conditions need to be designed for each domain, the proposed method switches between converters for layout representations, enabling the generation of a variety of scenes from indoor to outdoor.

\subsection{Scene Generation Using 360-Degree Image}

Image generation methods have been studied for 360-degree images that record the field of view in all directions from a single observer's viewpoint. Methods to generate 360-degree images from one or a few normal images \cite{IndoorIllum,NeuralIllum,360ic,360ic_j,s2ic,omni_dreamer,hara2022sig-ss,wu2023ipoldm} and text prompts \cite{chen2022text2light,wang2023panodiff,Tang2023mvdiffusion} have been reported. Methods for panoramic three-dimensional structure prediction were also proposed \cite{Im2Pano3D,Lighthouse}.

Studies have also extended the observer space to generate 3D scenes with six degrees of freedom (DoF) from 360-degree RGB-D. In \cite{omninerf,hara2022enhance,360fusion-nerf,wang2023perf}, methods were proposed for constructing a 6-DoF 3D scene by training the NeRF from 360-degree RGB-D. LDM3D \cite{stan2023ldm3d} shows a series of pipelines that add channels of depth to the latent diffusion model (LDM) \cite{ldm}, generate 360-degree RGB-D from the text, and mesh it. Generating 3D scenes via 360-degree images is advantageous in terms of guaranteeing scene consistency and reducing computation. Our research attempts to generate 360-degree images from multiple conditions and 6-DoF 3D scenes by layout-conditioned depth estimation and training the NeRF.

\subsection{Monocular Depth Estimation}

Monocular depth estimation involves estimating the depth of each pixel in a single RGB image. In recent years, deep learning-based methods have progressed significantly, and methods based on convolutional neural networks \cite{roy2016depth,laina2016deeper,Kuznietsov_2017_CVPR,xu2017depth,xu2018structured,Wei2021leres,masoumian2022gcndepth} and transformers \cite{Farooq_Bhat_2021,Cheng2021swindepth,yang2021depth,SunSC21,Ranftl2022} have been proposed. Monocular depth estimation for 360-degree images was also investigated \cite{zioulis2018omnidepth,Garanderie2018EliminatingTB,marc2019panopopups,wang2019depth,zioulis2019spherical,BiFuse20,Pintore2021SDD,reyarea2021360monodepth,ai2023HRDFuse}. However, since the accuracy of conventional monocular depth estimation is not sufficient, this study aims to improve accuracy by combining it with layout conditions.

\section{Proposed Method}
\label{sec:method}

\putFigWWpdf{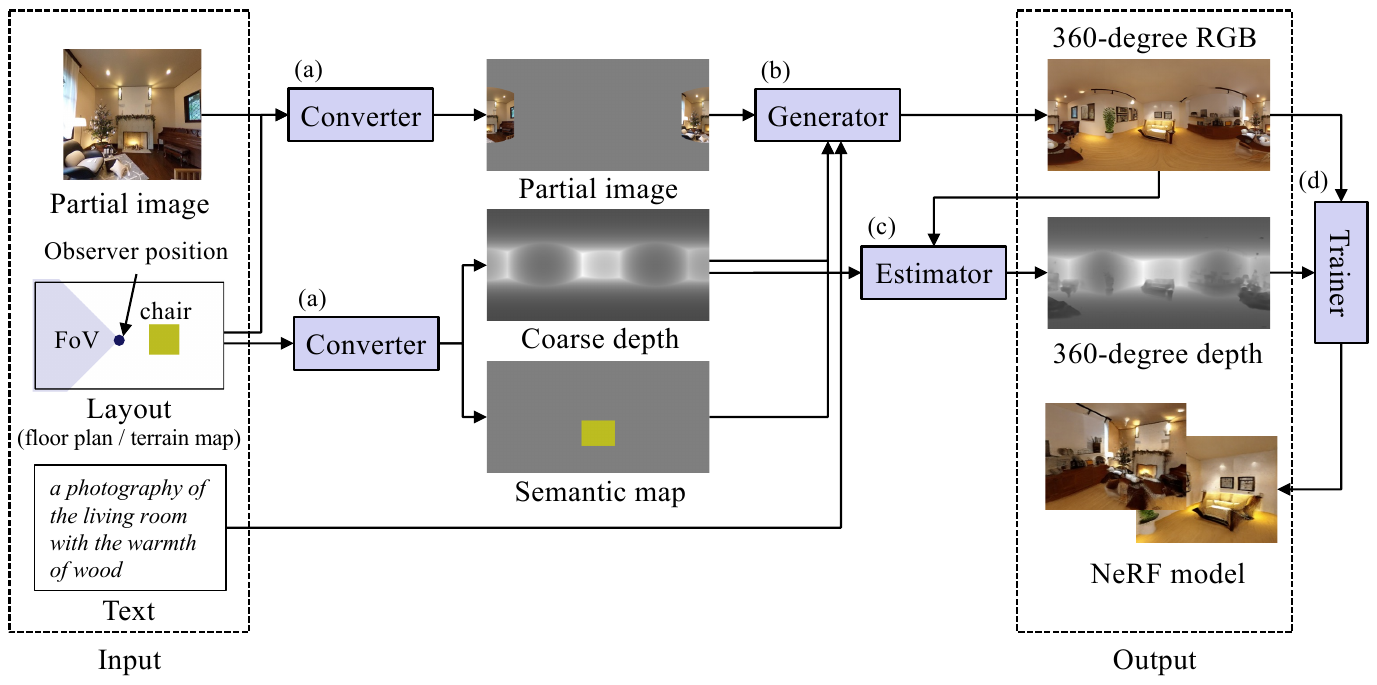}{Overview of the proposed method to generate 360-degree RGB-D and NeRF models from a partial image, layouts and text prompts.  (a) The partial image is converted to an ERP image from the observer position with the specified direction and field-of-view (FoV). The layout represented the in top view is converted to a coarse depth and a semantic map in ERP format with the observer position as the projection center. (b) These ERP images and texts are combined to generate a 360-degree RGB. (c) The generated RGB is combined with the coarse depth to estimate the fine depth. (d) a NeRF model is trained from 360-degree RGB-D.}{150mm}

This section describes the proposed method called \textit{MaGRITTe}, that generates 3D scenes under multiple conditions. \cref{fig:overview_3d} illustrates the overview of our method. Three input conditions are considered: a partial image, layout information represented in the top view, text prompts, and outputs from a 360-degree RGB-D and NeRF model. The proposed method comprises four steps: (a) ERP conversion of partial images and layouts, (b) 360-degree RGB image generation, (c) layout-conditioned depth estimation, and (d) NeRF training. The following sections describe each step.

\subsection{Conversion of Partial Image and Layout}
\label{sec:conversion}

First, we describe the conversion of the partial image and layout in (a) of \cref{fig:overview_3d}. This study uses two layout representations, floor plans and terrain maps, for indoor and outdoor scenes, respectively.

\subsubsection{Floor Plans}

A floor plan is a top-view representation of the room shape and  the position/size/class of objects. The room shape comprises the two-dimensional coordinates of the corners and the height positions of the floor and ceiling, based on the assumption that the walls stand vertically. The objects are specified by 2D bounding box, height from the floor at the top and bottom, and class, such as chair or table. 

\subsubsection{Terrain Maps}

\putFigWpdf{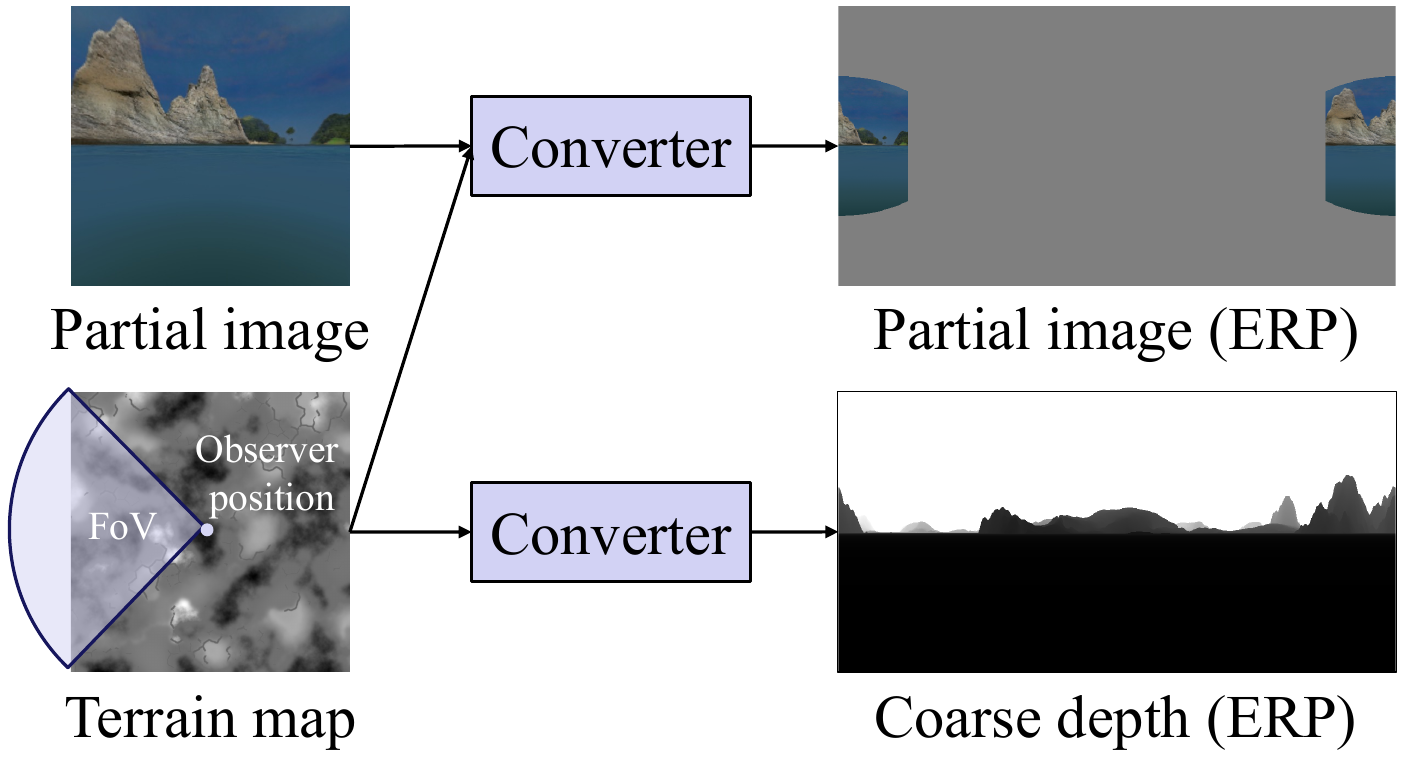}{The case of using a terrain map for the layout format. The partial image and the terrain map are converted into ERP images from the observer's viewpoint, respectively.}{78mm}

As shown in \cref{fig:terrain_map}, a terrain map describes the height of the terrain relative to the horizontal plane. This is a set $\mathbb{R}^{H_{\rm{ter}} \times W_{\rm{ter}}}$ that constitutes a $H_{\rm{ter}} \times W_{\rm{ter}}$ grid with the height of the ground surface at each grid point.

\subsubsection{ERP Conversion}

The observer position and field of view (FoV) of the partial image are provided in the layout. Based on this information, a partial RGB $\mathcal{P} \in \mathbb{R}^{H_{\rm{ERP}} \times W_{\rm{ERP}} \times 3}$, coarse depth $\mathcal{D} \in \mathbb{R}^{H_{\rm{ERP}} \times W_{\rm{ERP}}}$, and semantic map $\mathcal{S} \in \{0, 1\}^{H_{\rm{ERP}} \times W_{\rm{ERP}} \times C}$ are created in the ERP format, as shown in \cref{fig:overview_3d} (a), where $H_{\rm{ERP}}$ and  $W_{\rm{ERP}}$ are the height and width of the ERP image, respectively, and $C$ denotes the number of classes. The semantic map takes $\mathcal{S}_{ijc}=1$ when an object of class $c$ exists at position $(i, j)$ and $\mathcal{S}_{ijc}=0$ otherwise. For floor plans, the distance from the observer's viewpoint to the room wall is recorded, and for terrain maps, the distance from the observer's viewpoint to the terrain surface is recorded in ERP format and used as the coarse depth. A semantic map is created for a floor plan; the regions specifying the objects are projected onto the ERP image with the observer position of the partial image as the projection center, and object classes are assigned to the locations of their presence.

\subsection{360-Degree RGB Generation}

\putFigWpdf{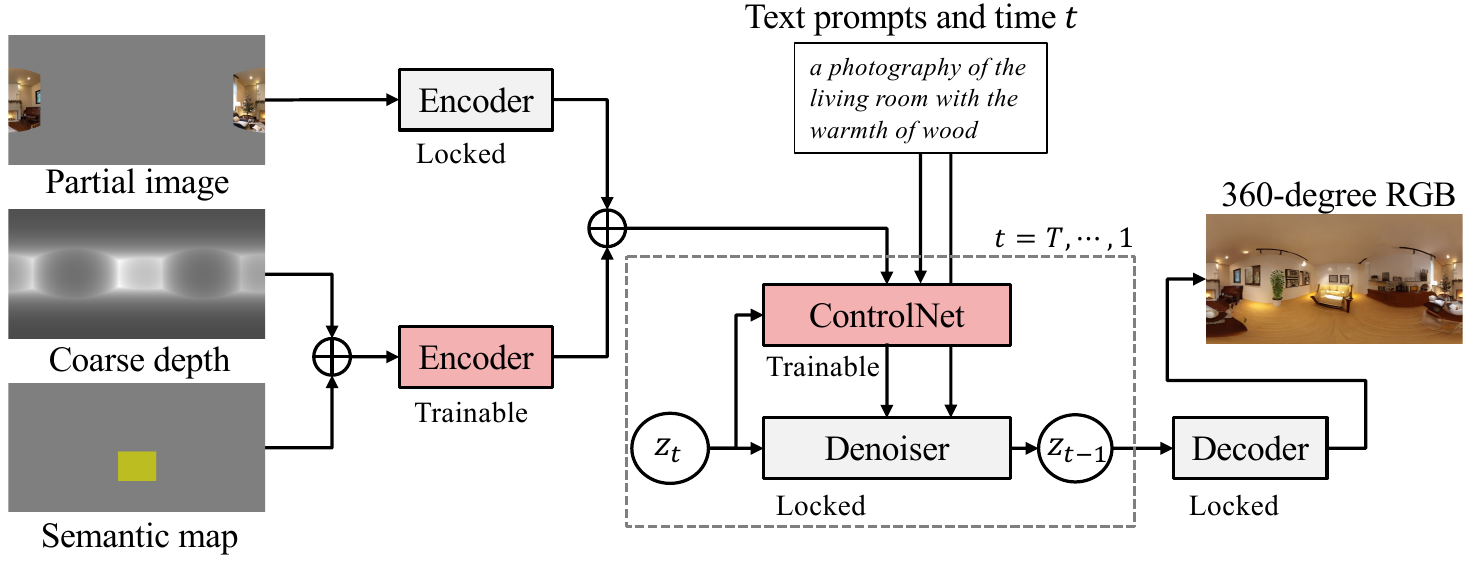}{The pipeline of generating 360-degree RGB from a partial image, coarse depth map, semantic map, and text prompts.}{144mm}

We combine partial images, coarse depths, and semantic maps represented in the ERP format and integrate them with text prompts to generate a 360-degree RGB image. Using the ERP format for the input and output allows the use of text-to-image models trained on large datasets. In this study, we employ StableDiffusion (SD) \cite{ldm}, a pre-trained diffusion model with an encoder and decoder, as the base text-to-image model. We fine-tune the model for our purposes using ControlNet \cite{zhang2023adding}, which controls the diffusion model with an additional network of conditional inputs. \cref{fig:generator} shows the pipeline to generate 360-degree RGB. A partial image, coarse depth, and semantic maps are embedded in the latent space, channel merged, and provided as conditional inputs to ControlNet along with text prompts. This is an improvement on PanoDiff \cite{wang2023panodiff}, which generates 360-degree images from partial images, and our method embeds layout information into a common latent space in ERP format as well, allowing for interaction between conditions while preserving spatial information. The encoder for partial images is from SD, and the encoder for layout information is a network with the same structure as that used in ControlNet. The weights of the network derived from SD are fixed, and only the weights of the network derived from ControlNet are updated during training.

\subsection{Layout-Conditioned Depth Estimation}
\label{sec:depth_refine}

Next, a fine depth is estimated from the coarse depth and the generated 360-degree RGB. In this study, we propose and compare two methods: end-to-end estimation and depth integration.

\subsubsection{End-to-End Estimation}

In the end-to-end approach, the depth is estimated using U-Net \cite{u-net} with a self-attention mechanism \cite{transformer} with four channels of RGB-D as the input, and one channel of depth as the output. The network is trained to minimize the L1 loss between the network outputs and ground truth. Details of the network configuration are provided in \cref{sec:depth_configs}.

\subsubsection{Depth Integration}

In the depth integration approach, depth estimates are obtained from 360-degree RGB using the monocular depth estimation method, LeRes \cite{Wei2021leres} is employed in this study, and the final depth is obtained so as to minimize the weighted squared error for the coarse depth and depth estimates. Since LeRes is designed for normal field-of-view images, the 360-degree image is projected onto $N$ tangent images, and depth estimation and integration are performed on each tangent image. Let $\hat{d_n} \in \mathbb{R}^{H_{\rm{d}}W_{\rm{d}}} (n=1, 2, \cdots, N)$ be the monocular depth estimate for $n$-th tangent image in ERP format, where $H_{\rm{d}}$ and $W_{\rm{d}}$ are the height and width of the depth map, respectively. Since the estimated depth $\hat{d}_n$ has unknown scale and offset, it is transformed using the affine transformation coefficient $s_n \in \mathbb{R}^2$ as $\tilde{d}_n s_n$, where $\tilde{d}_n = \begin{matrix} (\hat{d}_n & 1) \end{matrix} \in \mathbb{R}^{H_{\rm{d}}W_{\rm{d}} \times 2}$. We consider the following evaluation function $\mathcal{L}_{\rm{depth}}$, where $d_0 \in \mathbb{R}^{H_{\rm{d}}W_{\rm{d}}}$ is the coarse depth, $\Phi_n \in \mathbb{R}^{H_{\rm{d}}W_{\rm{d}} \times H_{\rm{d}}W_{\rm{d}}} (n=0, 1, \cdots, N)$ is the weight matrix, and $x \in \mathbb{R}^{H_{\rm{d}}W_{\rm{d}}}$ is the integrated depth.
\begin{equation}
\label{eq:depth_loss}
\mathcal{L}_{\rm{depth}} = || x - d_0 ||^2_{\Phi_0} + \sum_{n=1}^N || x - \tilde{d}_n s_n ||^2_{\Phi_n},
\end{equation}
where quadratic form $||v||^2_Q = v^\top Q v$. The fine depth $x$ and coefficients $s_n (n=1, 2, \cdots, N)$ that minimize $\mathcal{L}_{\rm{depth}}$ can be obtained in closed form from the extreme value conditions as follows:
\begin{equation}
\label{eq:x_cond}
x = \left( \sum_{n=0}^N  \Phi_n \right)^{-1} \left( \Phi_0 d_0 + \sum_{n=1}^N  \Phi_n  \tilde{d}_n s_n \right),
\end{equation}
\begin{equation}
\label{eq:s_cond2}
\begin{bmatrix}
    s_1 \\
    s_2  \\
    \vdots  \\
    s_N  \\
\end{bmatrix}
=
\begin{bmatrix}
    D_1 & U_{1,2} & \cdots & U_{1,N} \\
    U_{2,1} & D_2 & \cdots & U_{2,N} \\
    \vdots & \vdots & \ddots & \vdots \\
    U_{N,1} & U_{N,2}& \cdots& D_N \\
\end{bmatrix}^{-1}
\begin{bmatrix}
   b_1 \\
   b_2  \\
    \vdots  \\
   b_N  \\
\end{bmatrix},
\end{equation}
where, $D_k$ $=$ $\tilde{d}_k^{\top} \{ \Phi_k^{-1} + ( \sum_{n=0 \backslash k}^N  \Phi_n )^{-1}  \}^{-1}  \tilde{d}_k$, $U_{k,l}$ $=$ $- \tilde{d}_k^{\top} \Phi_k ( \sum_{n=0}^N  \Phi_n )^{-1}  \Phi_l  \tilde{d}_l, b_k = \tilde{d}_k^{\top} \Phi_k ( \sum_{n=0}^N  \Phi_n )^{-1} \Phi_0 d_0$\normalsize.
The derivation of the equation and setting of weights $\{\Phi_n\}_{n=0}^N$ are described in \cref{sec:derivation_integration}.

\subsection{Training NeRF}

Finally, we train the NeRF model using the generated 360-degree RGB-D. In this study, we employ a method from \cite{hara2022enhance} that can train NeRF by inpainting the occluded regions from a single image.

\section{Dataset}
\label{sec:data_creation}

We fine-tune our model using the following two types of datasets for indoor and outdoor scenes, respectively. We create artificial datasets with layout annotations using computer graphics as the \textit{base dataset}, whereas datasets without layout annotations are created using actual captured datasets as the \textit{auxiliary dataset}. 

\putFigWpdf{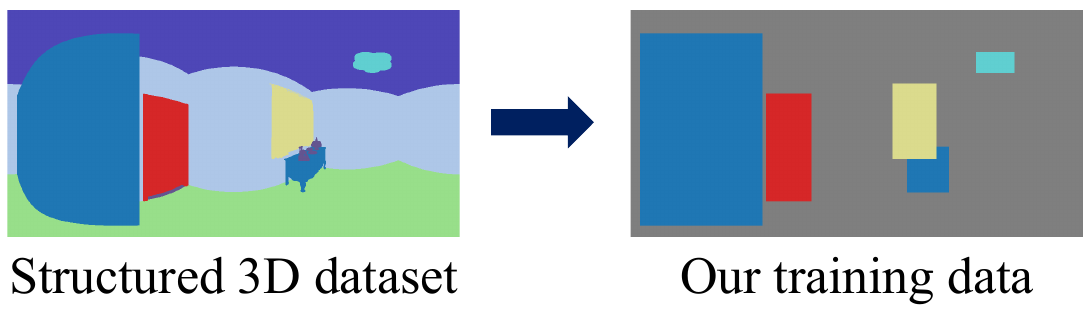}{Semantic map. Regions related to objects are extracted, excluding regions derived from the shape of the room, such as walls, floor, and ceiling, which are enclosed in a bounding box to form a semantic map in the proposed method.}{72mm}

\subsection{Indoor Scene}

For the base dataset, we modified and used a structured 3D dataset \cite{str3d} containing 3500 synthetic departments (scenes) with 185,985 panoramic renderings for RGB, depth, and semantic maps. The same room had both furnished and unfurnished patterns, and the depth of the unfurnished room was used as the coarse depth. For consistency with the ERP conversion in \cref{sec:conversion}, the semantic map was transformed, as shown in (\cref{fig:semantics}). Each image was annotated with text using BLIP \cite{li2022blip} and partial images were created using a perspective projection transformation of 360-degree RGB with random camera parameters. The data were divided into 161,126 samples for training, 2048 samples for validation, and 2048 samples for testing.

For the auxiliary dataset, we used the Matterport 3D dataset \cite{mat3d}, which is an indoor real-world 360$^\circ$ dataset including 10,800 RGB-D panoramic images. Similar to the structured 3D dataset, partial images and text were annotated. The depth and semantic maps included in the dataset were not used, and zero was assigned as the default value for the coarse depth and semantic map during training. The data were divided into 7675 samples for training and 2174 samples for testing.

\putFigWpdf{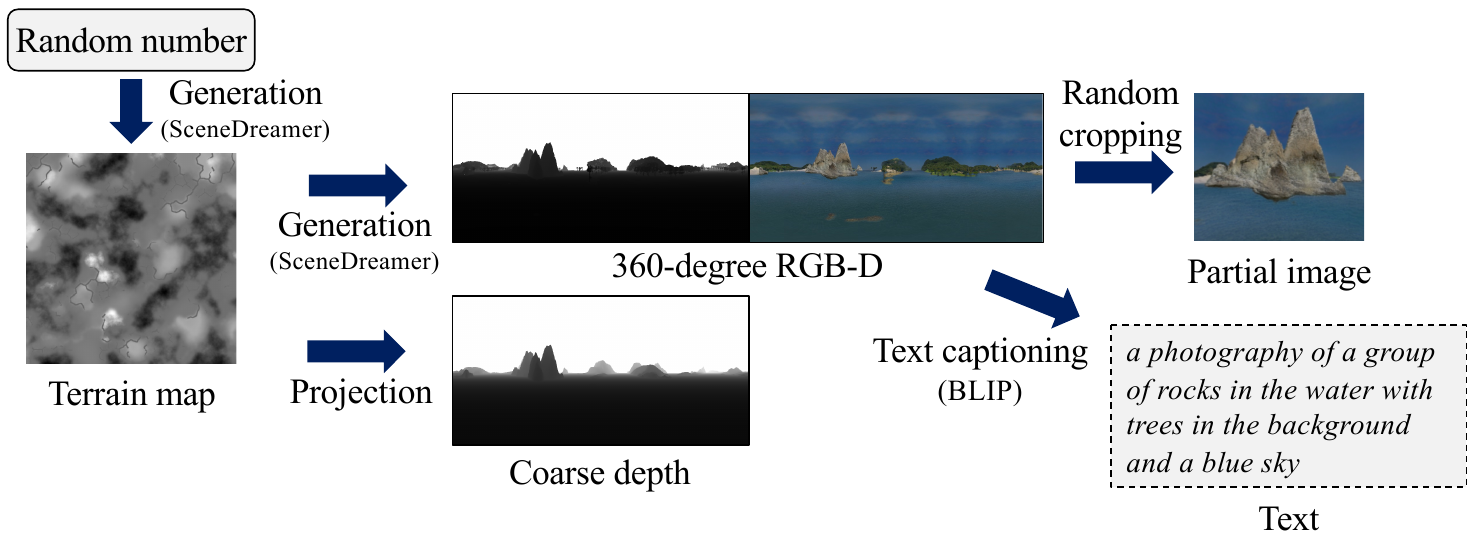}{Dataset creation for outdoor scene. SceneDreamer \cite{chen2023sd} generates a terrain map from a random number, and renders 360-degree RGB-D. The generated RGB image is annotated with text using BLIP \cite{li2022blip}, and partial images are created by a perspective projection transformation of 360-degree RGB with random camera parameters. A coarse depth is converted from the terrain maps}{144mm}

\subsection{Outdoor Scene}

As the base dataset, we created the \textit{SceneDreamer dataset} using SceneDreamer \cite{chen2023sd}, which is a model for generating 3D scenes. As shown in \cref{fig:outdoor_dataset}, a 360-degree RGB-D image was generated from random numbers via a terrain map to annotate the partial images and texts.  A semantic map was not used in this study because of limited object classes. The data were divided into 12,600 samples for training, 2,052 samples for validation, and 2052 samples for testing.

For the auxiliary dataset, we used the SUN360 dataset \cite{viewpoint} which includes various real captured 360-degree RGB images. We extracted only outdoor scenes from the dataset, and  partial images and text were annotated. The distance to the horizontal plane was set as the default value for the coarse depth during training. The data were divided into 39,174 training samples and 2048 testing samples.

\section{Experimental Results}
\label{sec:results}

Quantitative and qualitative experiments were conducted to verify the effectiveness of the proposed method, MaGRITTe, for generating 3D scenes under multiple conditions.

\subsection{Implementation Details}
\label{sec:impl}

The partial images, coarse depths, and semantic maps were in ERP format with a resolution of $512 \times 512$, and the shape of the latent variable in the LDM was $64 \times 64 \times 4$. We trained the 360-degree RGB generation model based on the pretrained SD v2.1 using the Adam optimizer \cite{adam} with a learning rate of $1.0 \times 10^{-5}$ and batch size of 16.  We trained the end-to-end depth estimation model from scratch using the Adam optimizer with a learning rate of $4.5 \times 10^{-6}$ and batch size of 6. The convolutional layers in the networks use circular padding \cite{hara2022sig-ss} to resolve the left-right discontinuity in ERP.

\subsection{360-Degree RGB Generation}

First, we evaluate 360-degree RGB generation. Because there is no comparison method that uses partial images, layouts, and text prompts as inputs to generate a 360-degree image, we compared our method with PanoDiff \cite{wang2023panodiff}, which is a state-of-the-art 360-degree RGB image generation model that uses partial images and texts. We implemented it and used PanoDiff with the encoder of the layout information removed in MaGRITTe for a fair comparison using the same network configurations and pretrained models.

\begin{table*}[tb]
\caption{Evaluation results of 360-degree RGB generation on the Modified Structured 3D dataset and the SceneDreamer dataset.}
\label{table:quant_rgb}
 \centering
 \scalebox{1.0}{
  \begin{tabular}{p{7.5em} | p{3.3em}   p{3.7em}  p{3.3em}  p{3.3em} | p{3.3em}  p{3.7em}  p{3.3em}  p{3.3em} } \hline
    \multicolumn{1}{c|}{} & \multicolumn{4}{c|}{Structured3D dataset} & \multicolumn{4}{c}{SceneDreamer dataset} \\ \hline
 \hfil method & \hfil  PSNR$\uparrow$ (whole)& \hfil PSNR$\uparrow$ (partial) & \hfil FID$\downarrow$ & \hfil CS$\uparrow$ & \hfil  PSNR$\uparrow$ (whole)& \hfil PSNR$\uparrow$ (partial) & \hfil 

FID$\downarrow$ & \hfil CS$\uparrow$ \\  \hline \hline
 \hfil PanoDiff \cite{wang2023panodiff} & \hfil 11.59 & \hfil \textbf{36.00} & \hfil  21.23 & \hfil \textbf{30.75} & \hfil 12.91 & \hfil \textbf{37.19} & \hfil  30.94 & \hfil 29.86 \\ 
 \hfil  MaGRITTe (ours)& \hfil  \textbf{12.56} & \hfil  35.39 & \hfil \textbf{18.87} & \hfil 30.72 & \hfil  \textbf{13.29} & \hfil  34.81 & \hfil \textbf{29.05} & \hfil \textbf{29.93} \\ \hline

 \end{tabular}
  }
\end{table*}

\begin{table}[b]
\caption{Evaluation results for object type and placement. Note that the object positions in the input condition are given by the bounding boxes as shown in \cref{fig:semantics}, therefore even in ground truth images, it doesn't match perfectly.}
\label{table:object}
 \centering
 \scalebox{1.0}{
  \begin{tabular}{p{8.2em}  | p{4em}  p{4em}   p{4em}} \hline
 \hfil Method & \hfil  Precision & \hfil Recall & \hfil IoU \\ \hline \hline
 \hfil Ground truth & \hfil 0.482 & \hfil 0.349 & \hfil 0.284  \\ \hline
 \hfil PanoDiff & \hfil 0.245 & \hfil 0.170& \hfil 0.124 \\ \hline
 \hfil MaGRITTe (ours) & \hfil 0.424 & \hfil 0.273 & \hfil 0.227 \\ \hline
 \end{tabular}
  }
\end{table}

\cref{table:quant_rgb} shows the quantitative evaluation results of 360-degree RGB generation on the Structured 3D dataset and the SceneDreamer dataset. We used the peak-signal-to-noise-ratio (PSNR) as the evaluation metric: PSNR (whole) for the entire image between the ground truth and generated images, PSNR (parial) for the region of the partial image given by the input. We also emply the FID \cite{fid}, which is a measure of the divergence of feature distributions between the ground truth and generated images, and the CLIP score (CS) \cite{radford2021learning,hessel-etal-2021-clipscore}, which promptly quantifies the similarity with the input text. PanoDiff is superior in terms of PSNR (partial) and CS, which is a reasonable result since PanoDiff is a method that takes only partial images and text prompts as conditions for image generation. However, MaGRITTe is superior to PSNR (whole) and FID, which indicates that the reproducibility and plausibility of the generated images can be enhanced by considering layout information as a condition as well. 

\cref{table:object} shows the results of the evaluation of the controllability of object type and placement. Semantic segmentation \cite{guerrero2020roomOR} was performed on the 360-degree images generated for Structured3D dataset to evaluate precision, recall, and IoU for bounding boxes in the input conditions. MaGRITTe is superior to PanoDiff and produces results closer to the ground truth images, indicating that the condition-aware object placement is realized.

\putFigWWpdf{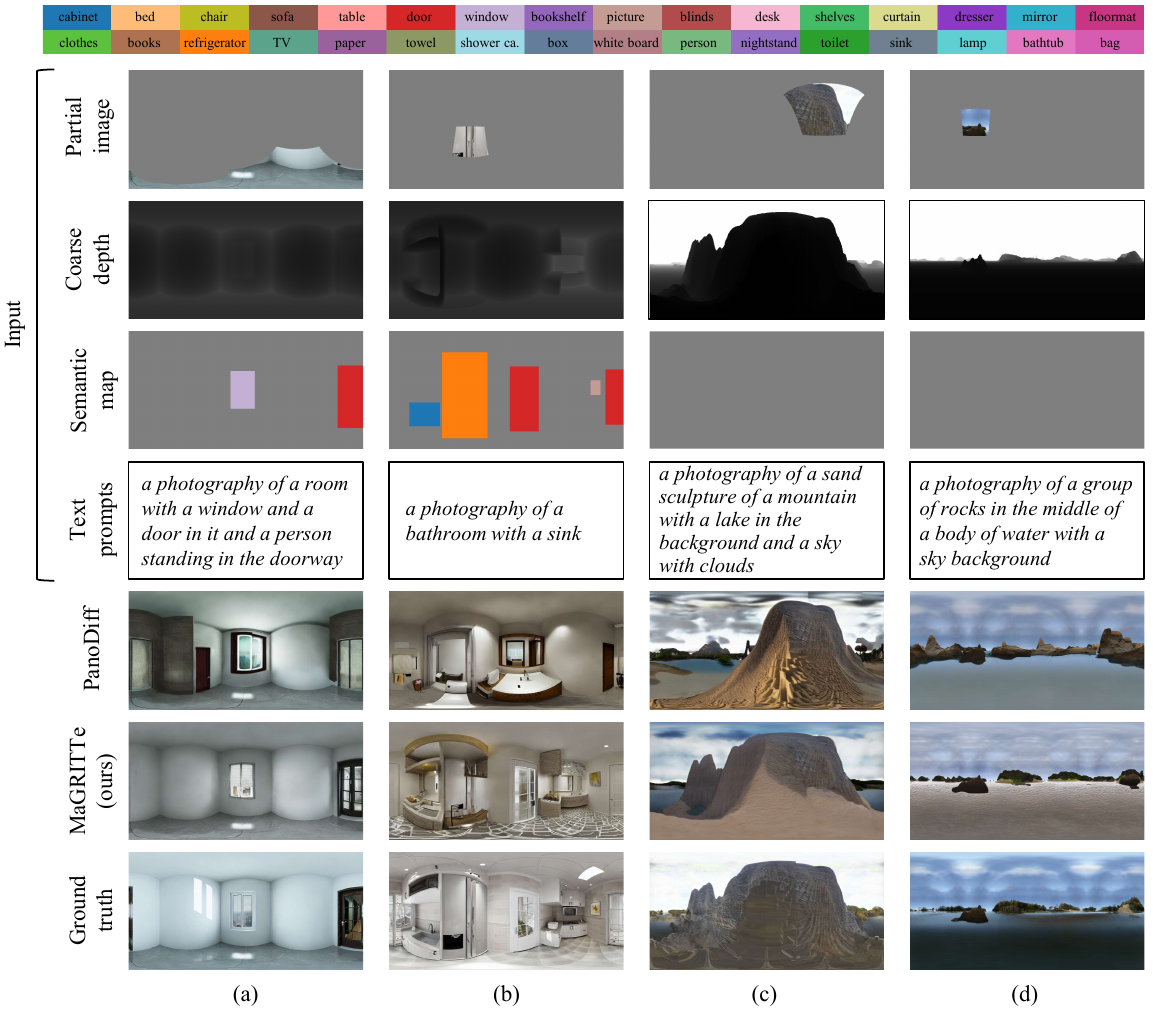}{The results of generating a 3D scene for the test set of (a)(b) the Stuructured 3D dataset and (c)(d) the SceneDreamer dataset.}{160mm}

\cref{fig:sample_360rgb} shows the examples of generating a 360-degree RGB image for the test set of the Structured 3D dataset and the SceneDreamer dataset. PanoDiff, which does not use the layout information as a condition,  generates images that differ significantly from the ground truth. This may have led to the degradation of PSNR (whole) and FID. Although the image generated by MaGRITTe differs from the ground-truth image at the pixel level, it can generate images with room geometry, terrain, and object placement in accordance with the given conditions. 

\subsection{360-Degree Depth Generation}

\begin{table}[tb]
\caption{Evaluation results of 360-degree depth generation on the Modified Structured 3D dataset and the SceneDreamer dataset}
\label{table:quant_depth}
 \centering
 \scalebox{1.0}{
  \begin{tabular}{p{9.8em} | p{4.0em}   p{4.0em}  | p{4.0em}   p{4.0em} } \hline
    \multicolumn{1}{c|}{} & \multicolumn{2}{c|}{Structured3D dataset} & \multicolumn{2}{c}{SceneDreamer dataset} \\ \hline
 \hfil Method & \hfil RMSE$\downarrow$ & \hfil AbsRel$\downarrow$ & \hfil RMSE$\downarrow$ & \hfil AbsRel$\downarrow$\\  \hline \hline
 Coarse depth  & \hfil 8.858 & \hfil 0.0117 & \hfil 15.30 & \hfil 0.0200 \\ \hline
 360MonoDepth \cite{reyarea2021360monodepth} & \hfil 21.67 & \hfil 0.0138 & \hfil 15.30 & \hfil 0.0202 \\ 
 LeRes (ERP) \cite{Wei2021leres} & \hfil 19.03 & \hfil 0.0149 & \hfil \underline{15.24} & \hfil \underline{0.0187} \\
 LeRes (multi views) & \hfil 21.90 & \hfil 0.0147 & \hfil 15.25 & \hfil 0.0188 \\ 
 Ours (end-to-end) & \hfil \textbf{6.649} & \hfil  \textbf{0.0056} & \hfil 15.29 & \hfil  0.0196 \\  
 Ours (depth integration) & \hfil \underline{7.432} & \hfil 0.0119 & \hfil \textbf{15.20} & \hfil \textbf{0.0185}  \\ 
 Ours (w/o coarse depth) & \hfil 9.837 & \hfil \underline{0.0070} & \hfil15.28 & \hfil 0.0196 \\  \hline
 \end{tabular}
  }
\end{table}

Next, we evaluate the depth of the generated 360-degree image. Because the estimated depth has scale and offset degrees of freedom, its value was determined to minimize the squared error with the ground-truth depth, similar to the method presented in \cite{Ranftl2022}. We used the root mean squared error (RMSE) and mean absolute value of the relative error,  ${\rm AbsRel} = \frac{1}{M}\sum_{i=1}^M \frac{|z_i - z_i^*|}{z_i^*}$, where $M$ is the number of pixels, $z_i$ is the estimated depth of the $i$th pixel, and $z_i^*$ is the ground-truth depth of the $i$th pixel. Pixels at infinity were excluded from evaluation. \cref{table:quant_depth} shows the results of the quantitative evaluation of depth generation on the Structured 3D dataset and the SceneDreamer dataset. For comparison, the results of 360MonoDepth which is a 360$^\circ$ monocular depth estimation \cite{reyarea2021360monodepth} method; LeRes (ERP), which is LeRes \cite{Wei2021leres} directly applied to ERP; and LeRes (multi views), which applies LeRes to multiple tangent images of a 360-degree image and integrates the estimated depths in a \cref{sec:depth_refine} manner without using coarse depth, are also shown. In terms of RMSE and AbsRel, our method (end-to-end) was the best for the structured 3D dataset, and our method (depth integration) was the best for the SceneDreamer dataset. It was also shown that combining LeRes with coarse depth increased accuracy compared to using LeRes alone. Ours (w/o coarse depth) is an end-to-end depth estimation method that uses only RGB without the coarse depth, and we can see that the accuracy is lower than when using coarse depth in the Structured3D dataset. The end-to-end method is relatively ineffective for the SceneDreamer dataset. This may be because the number of samples in the dataset was small and the depth was estimated to be close to the coarse depth.

\subsection{Results in the Wild}

\putFigWWpdf{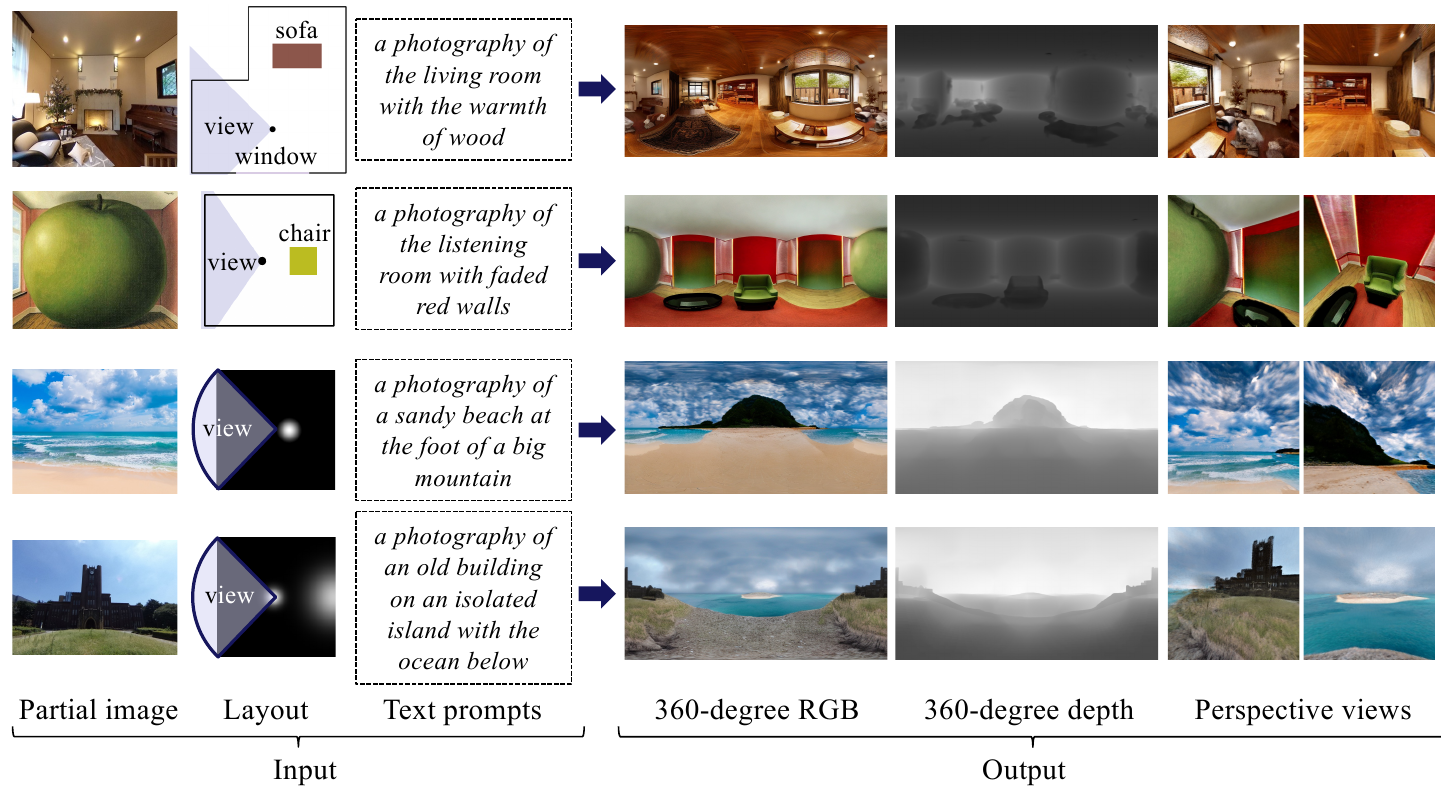}{Samples of the 3D scene generation based on user-generated conditions. Perspective views are rendered using the learned NeRF model. The first and fourth partial images are taken by the author using a camera, the second is a painting entitled "The Listening Room" by René Magritte and the third was downloaded from the web (https://www.photo-ac.com/).}{160mm}

We evaluated the results of 3D scene generation based on user-generated conditions outside the dataset used for fine-tuning. Examples of 3D scenes generated by MaGRITTe, conditioned on partial images, layouts, and text, are shown in \cref{fig:summary_360rgbd_gen,fig:sample_wild}. These conditions were created freely by the authors. It can be seen that the generated scene contains the given partial image and conforms to the instructions of the text prompt according to the given layout. These results show that MaGRITTe can generate 3D scenes with the appearance, geometry, and overall context controlled according to the input information, even outside the dataset used for fine-tuning. 

\subsection{Generation Results from Subset of Conditions}

\begin{table}[tb]
\caption{Evaluation results for generation from subset of conditions.}
\label{table:eval_subset}
 \centering
 \scalebox{1.0}{
  \begin{tabular}{p{2.5em}  p{2.2em}  p{2.2em}  | p{3.2em} p{3.3em} p{2.5em} p{2.5em}  | p{3.5em} p{3.8em} } \hline
  \multicolumn{3}{c|}{Conditions} & \multicolumn{4}{c|}{RGB} & \multicolumn{2}{c}{Depth} \\ \hline
 Partial image & Layout & Text &  \hfil  PSNR$\uparrow$ (whole)& \hfil PSNR$\uparrow$ (partial) & \hfil FID$\downarrow$ & \hfil CS$\uparrow$ & \hfil RMSE $\downarrow$ & \hfil AbsRel $\downarrow$ \\  \hline \hline
 \hfil $\checkmark$ & \hfil  $\checkmark$ & \hfil  $\checkmark$ & \hfil 12.42  & \hfil 33.29  & \hfil 18.84  & \hfil 30.71 & \hfil 5.05 & \hfil 0.0076 \\ \hline
 \hfil $\checkmark$ & \hfil $\checkmark$ & \hfil & \hfil  12.04  & \hfil 34.46  & \hfil 43.86  & \hfil 28.19 & \hfil 8.96 & \hfil 0.0100  \\ \hline
 \hfil & \hfil $\checkmark$ & \hfil  $\checkmark$ & \hfil 11.45 & \hfil -  & \hfil 21.71  & \hfil 30.67   & \hfil 8.78 & \hfil 0.0056 \\ \hline
 \hfil $\checkmark$ & \hfil & \hfil $\checkmark$ & \hfil  11.48  & \hfil 33.64  & \hfil 21.83  & \hfil 30.93  & \hfil 24.56 & \hfil 0.0172 \\ \hline
 \hfil $\checkmark$ & \hfil & \hfil & \hfil  11.40  & \hfil 35.00  & \hfil 55.08  & \hfil 27.00  & \hfil 23.94 & \hfil 0.0158 \\ \hline
 \hfil & \hfil $\checkmark$ & \hfil  & \hfil 11.12& \hfil -  & \hfil 59.70  & \hfil 27.59   & \hfil 5.02 & \hfil 0.0086 \\ \hline
 \hfil & \hfil & \hfil $\checkmark$ & \hfil  10.67  & \hfil -  & \hfil 25.90  & \hfil 30.85 & \hfil 24.53 & \hfil 0.0171 \\ \hline
 \hfil & \hfil & \hfil & \hfil  10.43  & \hfil -  & \hfil 87.69  & \hfil 24.40  & \hfil 24.00 & \hfil 0.0180 \\ \hline

 \end{tabular}
  }
\end{table}

To verify the contribution and robustness of each condition of the proposed method, experiments were conducted to generate 360-degree RGB-D from a subset of partial images, layouts, and text prompts. Generation was performed for the test set of the structured 3D dataset. Because depth estimation in MaGRITTe requires layout information, LeRes (ERP) \cite{Wei2021leres}, a monocular depth estimation of ERP images, was used in the absence of layout conditions. \cref{table:eval_subset} shows the values of each evaluation metric for the generated results. In terms of FID, it can be seen that MaGRITTe does not significantly degrade performance when text conditions are included in the generation conditions. This is largely owing to the performance of the text-to-image model used as the base model to ensure the plausibility of the generated image. However, PSNR (whole) decreases in the absence of partial image and layout conditions, indicating that the contribution of these conditions to the composition of the overall structure is high. In addition, CS naturally decreases without the text condition. However, even without the text condition, CS is larger than that in the unconditional generation case, indicating that semantic reproduction is possible to some extent, even from partial images and layout information. For depth generation, the accuracy is significantly degraded because it is impossible to use depth estimation with a coarse depth in the absence of layout conditions. When generated from partial images and text, its performance was comparable to PanoDiff. Details of the experimental setup, additional samples, ablation studies, and limitations are described in \cref{sec:additional,sec:discussion}.

\section{Conclusions}
\label{sec:conclusions}

We proposed a method for generating and controlling 3D scenes using partial images, layout information, and text prompts. We confirmed that fine-tuning a large-scale text-to-image model with small artificial datasets can generate 360-degree images from multiple conditions, and free perspective views can be generated by layout-conditioned depth estimation and training NeRF. This enables 3D scene generation from multimodal conditions without creating a new large dataset. It is also indicated that the interaction of multiple spatial conditions can be performed using a common ERP latent space, and that both indoor and outdoor scenes can be handled by replacing the conversions.

Future studies will include the detection of inconsistent input conditions and suggestions for users on how to resolve these inconsistencies. Creating conditions under which the layout and partial images match perfectly is difficult, and a method that aligns with the approximate settings is desirable.

\section*{Acknowledgements}

This work was partially supported by JST Moonshot R\&D Grant Number JPMJPS2011, CREST Grant Number JPMJCR2015 and Basic Research Grant (Super AI) of Institute for AI and Beyond of the University of Tokyo. We would like to thank Yusuke Kurose, Jingen Chou, Haruo Fujiwara, and Sota Oizumi for helpful discussions.


\bibliographystyle{splncs04}
\bibliography{references}

\clearpage

\appendix

\section{Details of Layout-Conditioned Depth Estimation}
\label{sec:details_depth}

\putFigWpdf{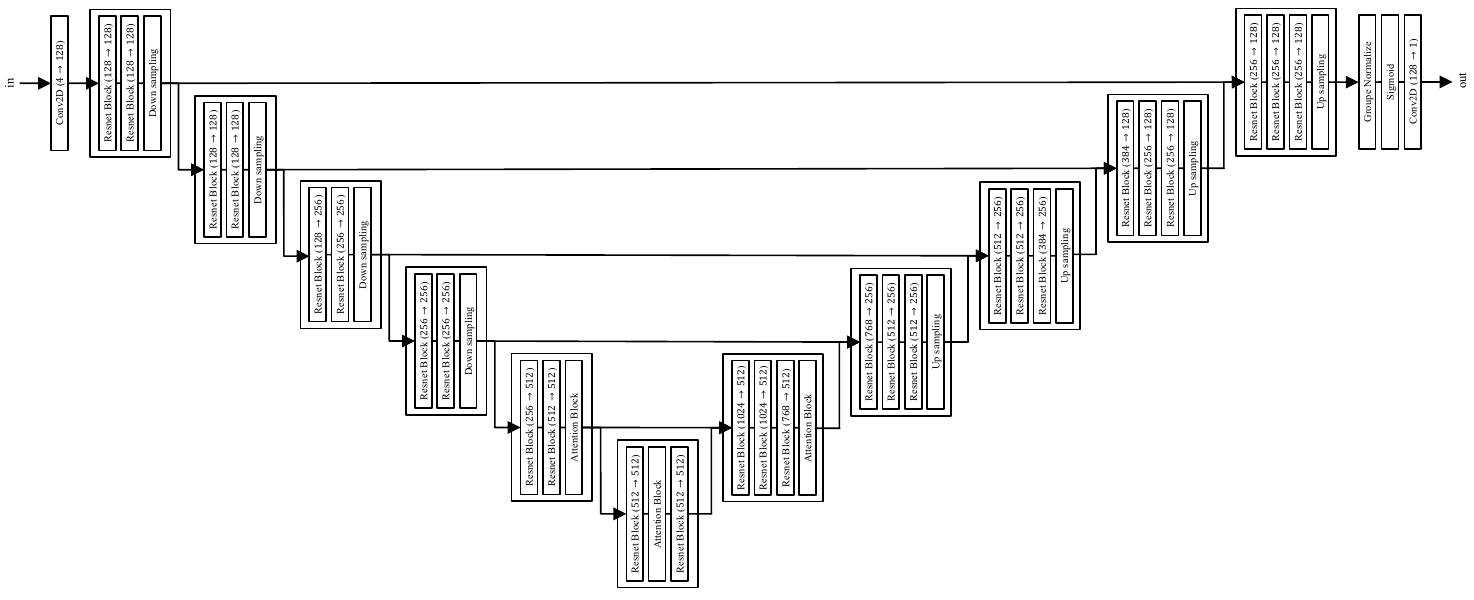}{The structure of the layout-conditioned depth estimation network. Conv2D ($N \to M$) is a two-dimensional convolutional layer with $N$ input channels, $M$ output channels, and a kernel size of $3 \times 3$. The Resnet Block shown in \cref{fig:resnet_block} is combined into a U-Net structure. Downsampling and upsampling are performed using a factor of 2. In the Attention Block, self-attention \cite{transformer} in the form of a query, key, and value is applied in pixels.}{160mm}

\putFigWpdf{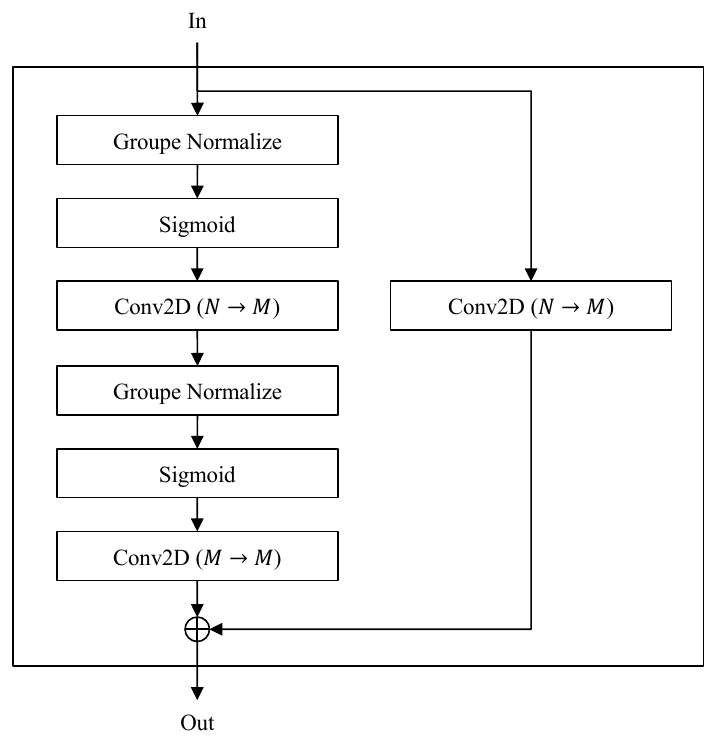}{The structure of a Resnet Block ($N \to M$). $N$ is the number of input channels, and $M$ is the number of output channels. In the groupe normalize, the number of split channels is fixed at 32. Conv2D refers to a two-dimensional convolutional layer, and the numbers in parentheses indicate the conversion of the number of channels.}{96mm}

In this section, we describe the details of the layout-conditioned depth estimation, which generates a fine depth from the coarse depth and generated RGB.

\subsection{End-toEnd Network Configuration}
\label{sec:depth_configs}

The structure of the network that generates a fine depth from a coarse depth and the generated RGB end-to-end is shown in \cref{fig:depth_net,fig:resnet_block}. The network consists of a combination of U-Net \cite{u-net} and self-attention \cite{transformer}, with four channels of RGB-D as the input and one channel of depth as the output. The network was trained to minimize the L1 loss between the depth output from the network and the depth of the ground truth. The model was trained from scratch using the Adam optimizer with a learning rate of $4.5 \times 10^{-6}$ and a batch size of six.

\subsection{Equation Derivation for Depth Integration}
\label{sec:derivation_integration}

Let $\hat{d_n} \in \mathbb{R}^{H_{\rm{d}}W_{\rm{d}}} (n=1, 2, \cdots, N)$ be the monocular depth estimate for $n$-th tangent image in ERP format, where $H_{\rm{d}}$ and $W_{\rm{d}}$ are the height and width of the depth map, respectively. Since the estimated depth $\hat{d}_n$ has unknown scale and offset, it is transformed using the affine transformation coefficient $s_n \in \mathbb{R}^2$ as $\tilde{d}_n s_n$, where $\tilde{d}_n = \begin{matrix} (\hat{d}_n & 1) \end{matrix} \in \mathbb{R}^{H_{\rm{d}}W_{\rm{d}} \times 2}$. We consider the following evaluation function $\mathcal{L}_{\rm{depth}}$, where $d_0 \in \mathbb{R}^{H_{\rm{d}}W_{\rm{d}}}$ is the coarse depth, $\Phi_n \in \mathbb{R}^{H_{\rm{d}}W_{\rm{d}} \times H_{\rm{d}}W_{\rm{d}}} (n=0, 1, \cdots, N)$ is the weight matrix, and $x \in \mathbb{R}^{H_{\rm{d}}W_{\rm{d}}}$ is the integrated depth.
\begin{equation}
\label{eq:depth_loss}
\mathcal{L}_{\rm{depth}} = || x - d_0 ||^2_{\Phi_0} + \sum_{n=1}^N || x - \tilde{d}_n s_n ||^2_{\Phi_n},
\end{equation}
where the quadratic form $||v||^2_Q = v^\top Q v$. We find the affine transformation coefficient $s_n (n=1, 2, \cdots, N)$ and fine depth $x$ from the extreme-value conditions to minimize $\mathcal{L}_{\rm{depth}}$. The partial differentiation of \cref{eq:depth_loss} with $x$ yields:
\begin{align}
\label{eq:partial_x}
\frac{\partial \mathcal{L}_{\rm{depth}}}{\partial x} &= 2 \Phi_0 (x - d_0) + 2 \sum_{n=1}^N  \Phi_n (x- \tilde{d}_n s_n) \nonumber \\
&= 2 \sum_{n=0}^N  \Phi_n x - 2 \left( \Phi_0 d_0 +   \sum_{n=1}^N  \Phi_n  \tilde{d}_n s_n \right),
\end{align}
and $x$ satisfying the extreme-value conditions are as follows:
\begin{equation}
\label{eq:x_cond}
x = \left( \sum_{n=0}^N  \Phi_n \right)^{-1} \left( \Phi_0 d_0 + \sum_{n=1}^N  \Phi_n  \tilde{d}_n s_n \right).
\end{equation}
Next, the partial differentiation of \cref{eq:depth_loss} with $s_k$ yields:
\begin{equation}
\frac{\partial \mathcal{L}_{\rm{depth}}}{\partial s_k} = -2 \tilde{d}_k^{\top} \Phi_k (x- \tilde{d}_k s_k),
\end{equation}
and $s_k$ satisfying the extreme-value conditions are as follows:
\begin{equation}
\tilde{d}_k^{\top} \Phi_k \tilde{d}_k s_k = \tilde{d}_k^{\top} \Phi_k x.
\end{equation}
By substituting \cref{eq:x_cond} into \cref{eq:s_cond}, we obtain
\begin{equation}
\tilde{d}_k^{\top} \Phi_k \tilde{d}_k s_k = \tilde{d}_k^{\top} \Phi_k \left( \sum_{n=0}^N  \Phi_n \right)^{-1} \left( \Phi_0 d_0 + \sum_{n=1}^N  \Phi_n  \tilde{d}_n s_n \right).
\end{equation}
Transposing $s_n$ on the left-hand side yields
\begin{equation}
\label{eq:s_cond}
\tilde{d}_k^{\top} \Phi_k \tilde{d}_k s_k - \tilde{d}_k^{\top} \Phi_k \left( \sum_{n=0}^N  \Phi_n \right)^{-1}  \sum_{n=1}^N  \Phi_n  \tilde{d}_n s_n  = \tilde{d}_k^{\top} \Phi_k \left( \sum_{n=0}^N  \Phi_n \right)^{-1} \Phi_0 d_0.
\end{equation}

Considering the coefficient of $s_k$ as $D_k \in \mathbb{R}^{2 \times 2}$, we obtain
\begin{align}
\label{eq:d_k}
D_k &= \tilde{d}_k^{\top} \Phi_k \tilde{d}_k - \tilde{d}_k^{\top} \Phi_k \left( \sum_{n=0}^N  \Phi_n \right)^{-1}  \Phi_k  \tilde{d}_k  \nonumber \\
&= \tilde{d}_k^{\top} \Phi_k \left\{ I - \left( \sum_{n=0}^N  \Phi_n \right)^{-1}  \Phi_k \right\}  \tilde{d}_k \nonumber \\
&= \tilde{d}_k^{\top} \Phi_k \left\{ I - \left( I +  \Phi_k^{-1} \sum_{n=0 }^{N\backslash k}  \Phi_n \right)^{-1} \right\}  \tilde{d}_k \nonumber \\
&= \tilde{d}_k^{\top} \Phi_k \left\{ I + \left( \sum_{n=0 }^{N\backslash k}  \Phi_n \right)^{-1} \Phi_k  \right\}^{-1}  \tilde{d}_k \nonumber \\
&= \tilde{d}_k^{\top} \left\{ \Phi_k^{-1} + \left( \sum_{n=0 }^{N\backslash k}  \Phi_n \right)^{-1}  \right\}^{-1}  \tilde{d}_k,
\end{align}
where $\sum_{n=0 }^{N\backslash k} \Phi_n := \sum_{n=0}^N \Phi_n - \Phi_k$.
In addition, considering the coefficient of $s_l (l \neq k)$ as $U_{k,l} \in \mathbb{R}^{2 \times 2}$, we obtain
\begin{equation}
\label{eq:beta_coef}
U_{k,l} = - \tilde{d}_k^{\top} \Phi_k \left( \sum_{n=0}^N  \Phi_n \right)^{-1}  \Phi_l  \tilde{d}_l.
\end{equation}
The constant $b_k \in \mathbb{R}^2$ is expressed as follows:
\begin{equation}
\label{eq:gamma_coef}
b_k = \tilde{d}_k^{\top} \Phi_k \left( \sum_{n=0}^N  \Phi_n \right)^{-1} \Phi_0 d_0.
\end{equation}
Therefore, when the conditions in \cref{eq:s_cond} are coupled for $k=1, 2, \cdots, N$, we obtain
\begin{equation}
\begin{bmatrix}
    D_1 & U_{1,2} & \cdots & U_{1,N} \\
    U_{2,1} & D_2 & \cdots & U_{2,N} \\
    \vdots & \vdots & \ddots & \vdots \\
    U_{N,1} & U_{N,2}& \cdots& D_N \\
\end{bmatrix}
\begin{bmatrix}
    s_1 \\
    s_2  \\
    \vdots  \\
    s_N  \\
\end{bmatrix}
=
\begin{bmatrix}
   b_1 \\
   b_2  \\
    \vdots  \\
   b_N  \\
\end{bmatrix}.
\end{equation}
We can then solve for $s_n (n=1, 2, \cdots, N)$  as follows.
\begin{equation}
\label{eq:s_cond2}
\begin{bmatrix}
    s_1 \\
    s_2  \\
    \vdots  \\
    s_N  \\
\end{bmatrix}
=
\begin{bmatrix}
    D_1 & U_{1,2} & \cdots & U_{1,N} \\
    U_{2,1} & D_2 & \cdots & U_{2,N} \\
    \vdots & \vdots & \ddots & \vdots \\
    U_{N,1} & U_{N,2}& \cdots& D_N \\
\end{bmatrix}^{-1}
\begin{bmatrix}
   b_1 \\
   b_2  \\
    \vdots  \\
   b_N  \\
\end{bmatrix}.
\end{equation}
From the above results, we can determine $x$ that minimizes equation \cref{eq:depth_loss} by first calculating $s_n (n=1, 2, \cdots, N)$ using \cref{eq:s_cond2} and then substituting the value into \cref{eq:x_cond}.

\subsection{Weight Setting for Depth Integration}

In this study, we set the weight matrix $\Phi_n  (n=0, 1, \cdots, N)$ to a diagonal matrix. By making it a diagonal matrix, the large matrix calculation in \cref{eq:d_k,eq:beta_coef,eq:gamma_coef} can be avoided and can be attributed to element-by-element calculations. The diagonal components represent the reflected intensity at each location on each depth map. Since the weight matrices $\Phi_n  (n=1, 2, \cdots, N)$ are for depth maps that express the estimated depth for $N$ tangent images in ERP format, the weights are increased for regions where tangent images are present, as shown in \cref{fig:weight_depth} To smooth the boundary, we first set the following weights $w_{ij}$ for pixel position $(i, j)$ in the tangent image of height $H_{\rm{tan}}$ and width $W_{\rm{tan}}$.
\begin{equation}
w_{ij} = \left\{ 1 -  \left( \frac{2i}{H_{\rm{tan}}} - 1 \right)^2 \right\}  \left\{ 1 -  \left( \frac{2j}{W_{\rm{tan}}} - 1 \right)^2 \right\}.
\end{equation}
This weight has a maximum value of 1 at the center of the tangent image and a minimum value of 0 at the edges of the image. The weights for the tangent image are converted to ERP format  and set to the diagonal components of the weight matrix $\Phi_n (n=1, 2, \cdots, N)$. The weights of the outer regions of each tangential image are set to zero. Tangent images are created with a horizontal field of view of 90 degrees and resolution of $512 \times 512$ pixels, and 16 images were created with the following latitude $\theta_n$ and longitude $\phi_n$ shooting directions.
\begin{equation}
\theta_n = \left\{
\begin{array}{ll}
\frac{\pi}{4} & (1 \leq n \leq 4)\\
-\frac{\pi}{4} & (5 \leq n \leq 8)\\
0 & (9 \leq n \leq 16)\\
\end{array}
\right.
\end{equation}
\begin{equation}
\phi_n = \left\{
\begin{array}{ll}
\frac{\pi n}{2} & (1 \leq n \leq 8)\\
\frac{\pi n}{4} & (9 \leq n \leq 16)
\end{array}
\right.
\end{equation}

On the other hand, the weights for the coarse depth $\Phi_0$ are set as follows. When using floor plans for the layout format, a low-weight $\eta_L$ is set for areas in the partial image or layout condition where an object is specified, and a high-weight $\eta_H (\geq \eta_L)$ for other areas. In this study, we set $\eta_L=0.0$, $\eta_H=2.0$. When using the terrain map for the layout format, set the diagonal component of the weight matrix $\Phi_0(i, j)$ according to the value of the coarse depth at each location $(i, j)$ in the ERP as follows:
\begin{equation}
\Phi_0(i, j) = \frac{\alpha }{d_0(i, j)^2 + \epsilon},
\end{equation}
where $\alpha$ and $\epsilon$ are hyperparameters. In this study, the coarse depth is normalized to the interval [0, 1], and we set $\alpha=1.0  \times 10^{-3}$ and $\epsilon=1.0 \times 10^{-8}$. We set $\Phi_0(i, j) = 0$ in the region where the coarse depth is infinite. The weights are inversely proportional to the square of the coarse depth to ensure that the squared error in \cref{eq:depth_loss} assumes values of the same scale with respect to the coarse depth. This prevents the error from being overestimated when an object is generated in the foreground of a large-depth region, such as a tree in the foreground of the sky.

\putFigWpdf{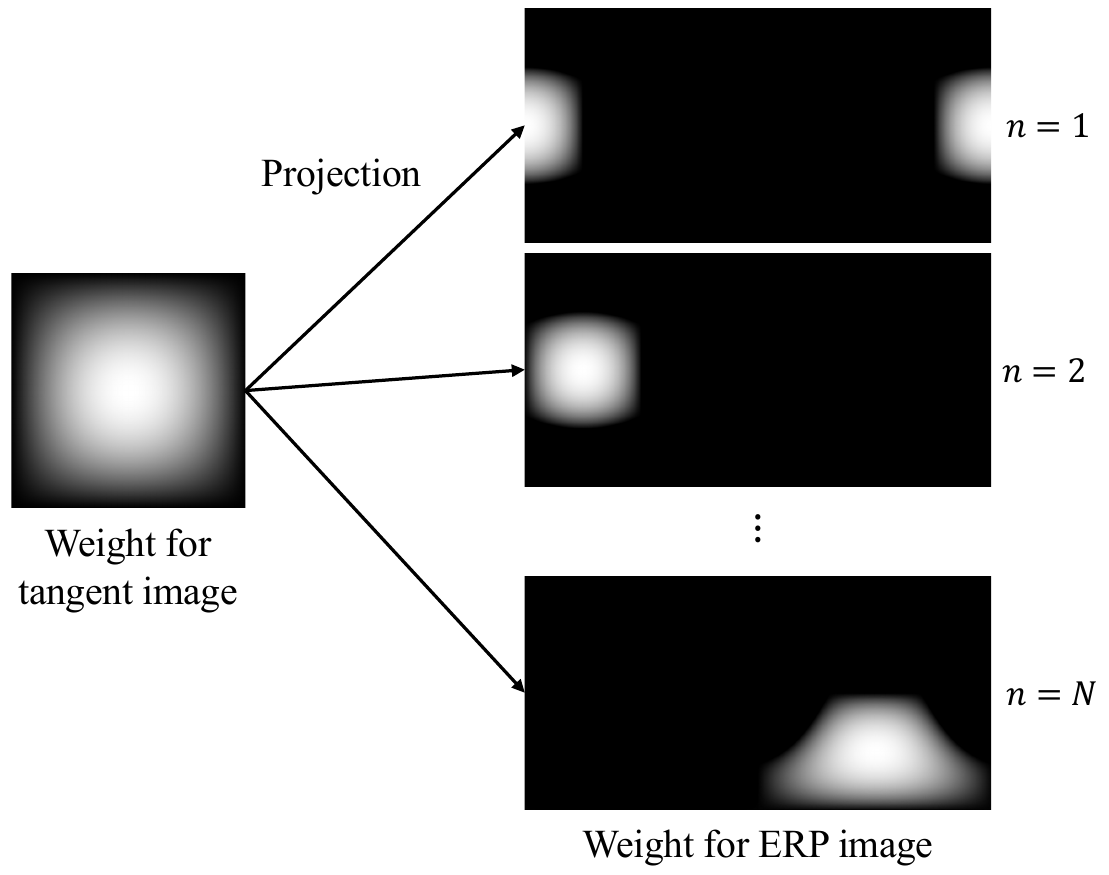}{Weights for estimated depth maps. The weights are set such that the center of the tangent image is 1, the edges of the image are 0, and  the weights are converted to ERP format for each depth map $ (n=1, 2, \cdots, N)$.}{80mm}

\section{Additional Results}
\label{sec:additional}

\subsection{Condition Dropout}

Fine-tuning of the base model degrades image-to-text performance. To mitigate this phenomenon, we additionally use the \textit{auxiliary dataset} (see Section 4) with text annotations only for fine-tuning. If one model is trained for different combinations of conditions, the learning may not be generalized to other combinations of conditions. We introduce condition dropout (CD), in which training is performed by randomly changing the combination of conditions. Each condition is dropped with a probability of 50\%, with the ERP image conditions being replaced by pixel values of 0 and text replaced by an empty string.

\cref{table:quant_cd} shows the results of comparing the presence or absence of CD in the proposed method. FID tended to be slightly better when CD was present, whereas PSNR (whole), PSNR (partial), and CS were superior or inferior depending on the two datasets. The better performance of CD on the SceneDreamer dataset can be attributed to the larger number of samples in the auxiliary dataset.

Next, we present the results of the evaluation of the experiment in a setting in which the conditions were crossed between datasets. \cref{table:quant_forgetting} shows the results of the CS for generated results with the text prompt of the auxiliary dataset for the depth of the base dataset. This indicates that CS can be improved by using the auxiliary dataset and CD. \cref{fig:sample_dropout} shows the difference with and without CD. These results show that the use of CD better reflects text prompts, and the generalization of text prompts in combination with depth is possible.

\begin{table*}[tb]
\caption{Evaluation results of 360-degree RGB generation on the Modified Structured 3D dataset and the SceneDreamer dataset.}
\label{table:quant_cd}
 \centering
 \scalebox{1.0}{
  \begin{tabular}{p{7.5em} | p{3.3em}   p{3.7em}  p{3.3em}  p{3.3em} | p{3.3em}  p{3.7em}  p{3.3em}  p{3.3em} } \hline
    \multicolumn{1}{c|}{} & \multicolumn{4}{c|}{Structured3D dataset} & \multicolumn{4}{c}{SceneDreamer dataset} \\ \hline
 \hfil method & \hfil  PSNR$\uparrow$ (whole)& \hfil PSNR$\uparrow$ (partial) & \hfil FID$\downarrow$ & \hfil CS$\uparrow$ & \hfil  PSNR$\uparrow$ (whole)& \hfil PSNR$\uparrow$ (partial) & \hfil FID$\downarrow$ & \hfil CS$\uparrow$ \\  \hline \hline
 \hfil  w/o CD& \hfil  \textbf{12.56} & \hfil  \textbf{35.39} & \hfil 18.87 & \hfil \textbf{30.72} & \hfil  12.46 & \hfil  34.68 & \hfil 29.54 & \hfil 29.71 \\  
 \hfil  w/ CD& \hfil  12.42 & \hfil  33.29 & \hfil \textbf{18.84} & \hfil 30.71 & \hfil  \textbf{13.29} & \hfil  \textbf{34.81} & \hfil \textbf{29.05} & \hfil \textbf{29.93} \\ \hline

 \end{tabular}
  }
\end{table*}

\begin{table}[tb]
\caption{CS evaluation results for base model forgetting}
\label{table:quant_forgetting}
 \centering
 \scalebox{1.0}{
  \begin{tabular}{p{5.2em}  p{6.2em}  p{4.2em}  | p{3.2em}  | p{3.6em}} \hline
 \begin{tabular}{c}Trained on \\base dataset \end{tabular} & \begin{tabular}{c}Trained on \\auxiliary dataset \end{tabular} & \begin{tabular}{c}Condition \\dropout \end{tabular} & \hfil Indoor & \hfil Outdoor \\  \hline \hline
 \hfil $\checkmark$ & \hfil  & \hfil  & \hfil 29.48 & \hfil 24.75 \\ \hline
 \hfil $\checkmark$ & \hfil $\checkmark$ & \hfil & \hfil 29.34 & \hfil 26.24 \\ \hline
 \hfil $\checkmark$ & \hfil $\checkmark$ & \hfil $\checkmark$ & \hfil  \textbf{30.23} & \hfil  \textbf{29.26} \\ \hline
 \end{tabular}
  }
\end{table}

\putFigWWpdf{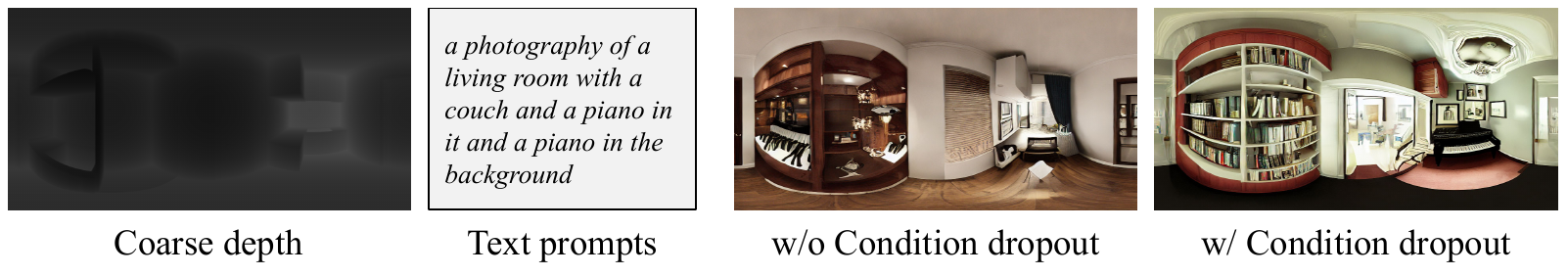}{The difference with and without CD. In this example, "piano" in the text prompt is reflected only for the method with CD.}{150mm}

\subsection{Comparison with Text2Room}

Text2Room \cite{hoellein2023text2room} is a method for generating 3D scenes as meshes by repeatedly generating images in multiple viewpoints from the input text. 
This method can also be used to control the layout of the generated 3D scene by changing the input text according to the viewpoint. 
However, layout guided generation in Text2Room is different from our setting, because it changes the text prompts for the direction of observation and cannot take geometric shapes as conditions. \cref{fig:text2room} shows an example of a scene generated by Text2Room under the same conditions as in \cref{fig:summary_360rgbd_gen}. Text2Room is less accurate in the placement of objects and is unable to generate room shapes to suit the conditions. Conditioning the layout with semantic map and coarse depth is the advantage of our method.

\putFigWWpdf{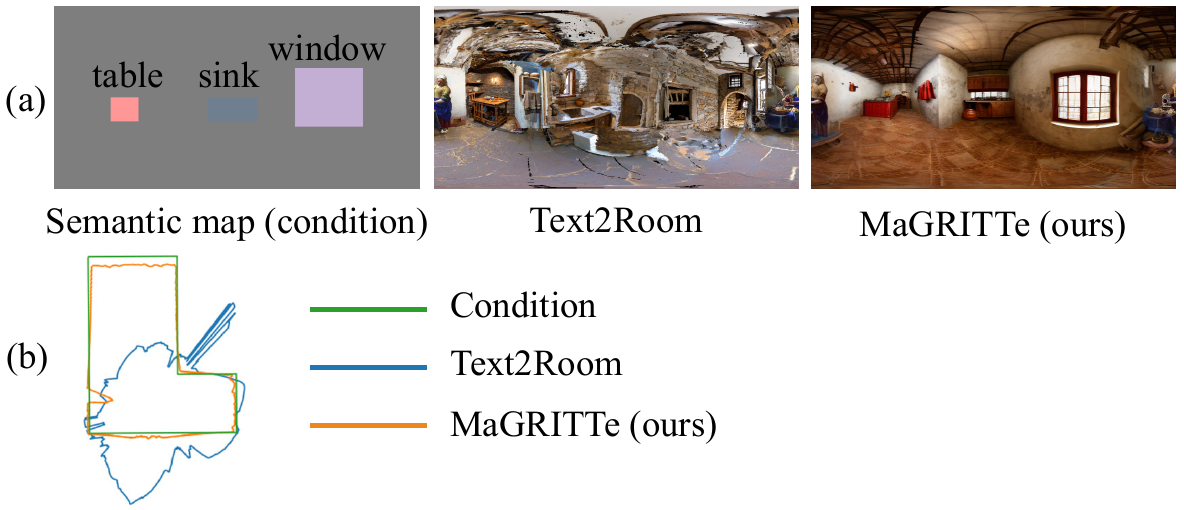}{Comparison with Text2Room. (a) ERP images of the generated 3D scenes, (b) Room shapes in the top view.}{100mm}

\subsection{360-Degree RGB Generation}

\putFigWWpdf{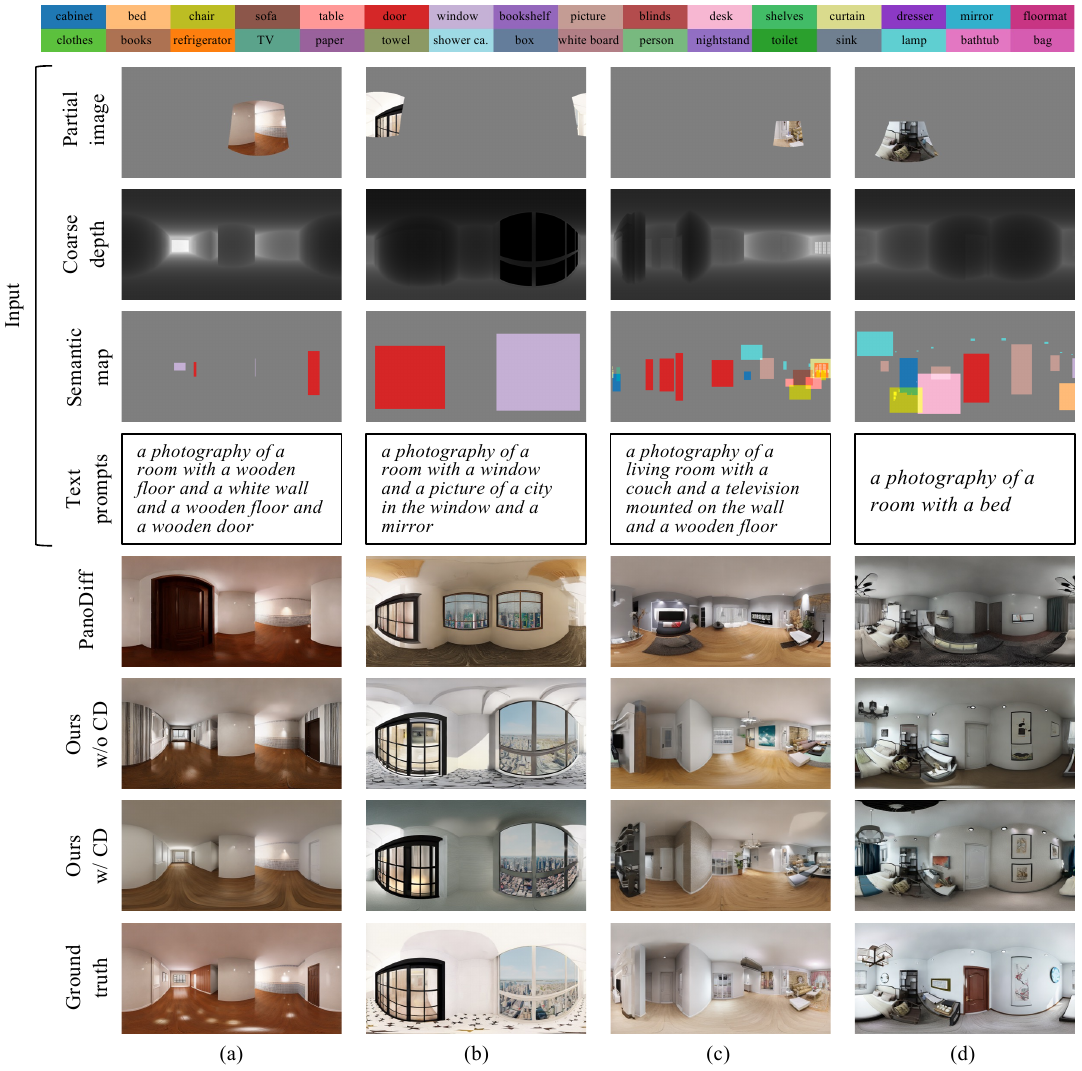}{The results of generating a 3D scene for the test set of the Stuructured 3D dataset.}{160mm}
\putFigWWpdf{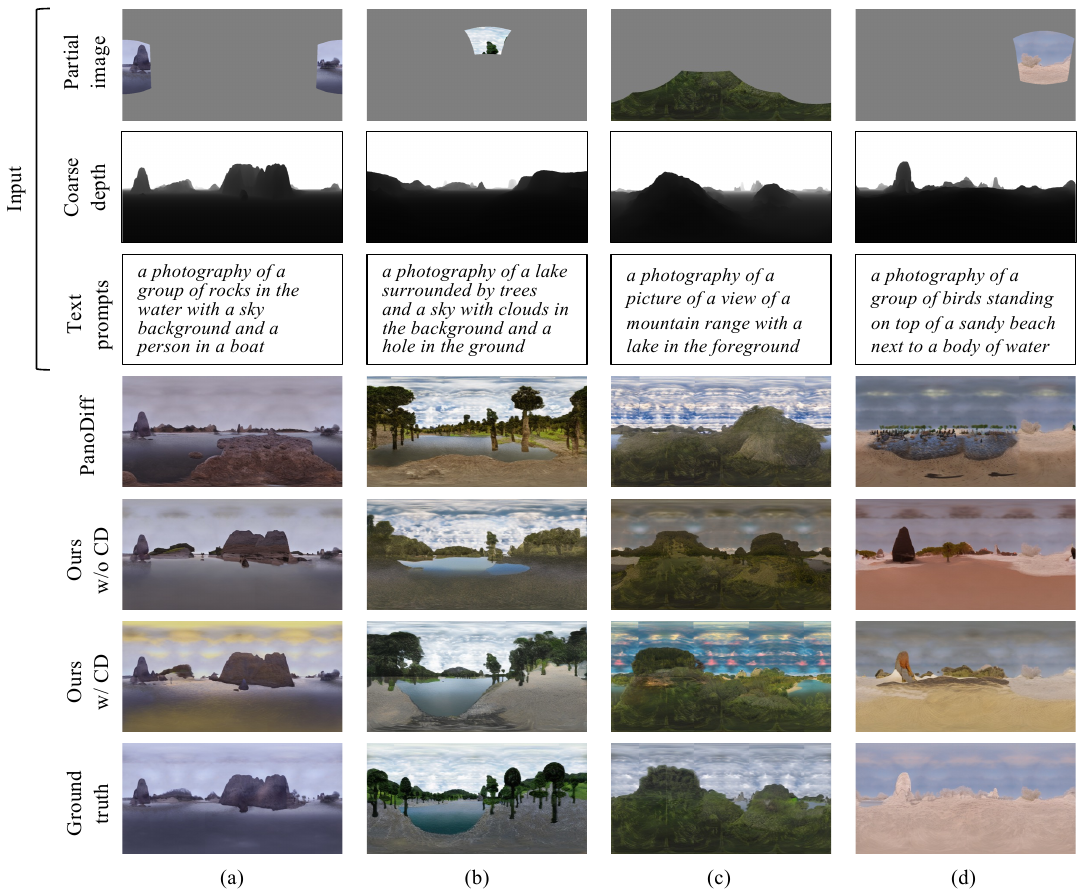}{The results of generating a 3D scene for the test set of the SceneDreamer dataset.}{160mm}

\cref{fig:sample_360rgb_str3d,fig:sample_360rgb_scdr} show additional samples of 360-degree RGB image generation for the Structured 3D dataset and SceneDreamer dataset, respectively.

\subsection{Results in the Wild}

\putFigWWpdf{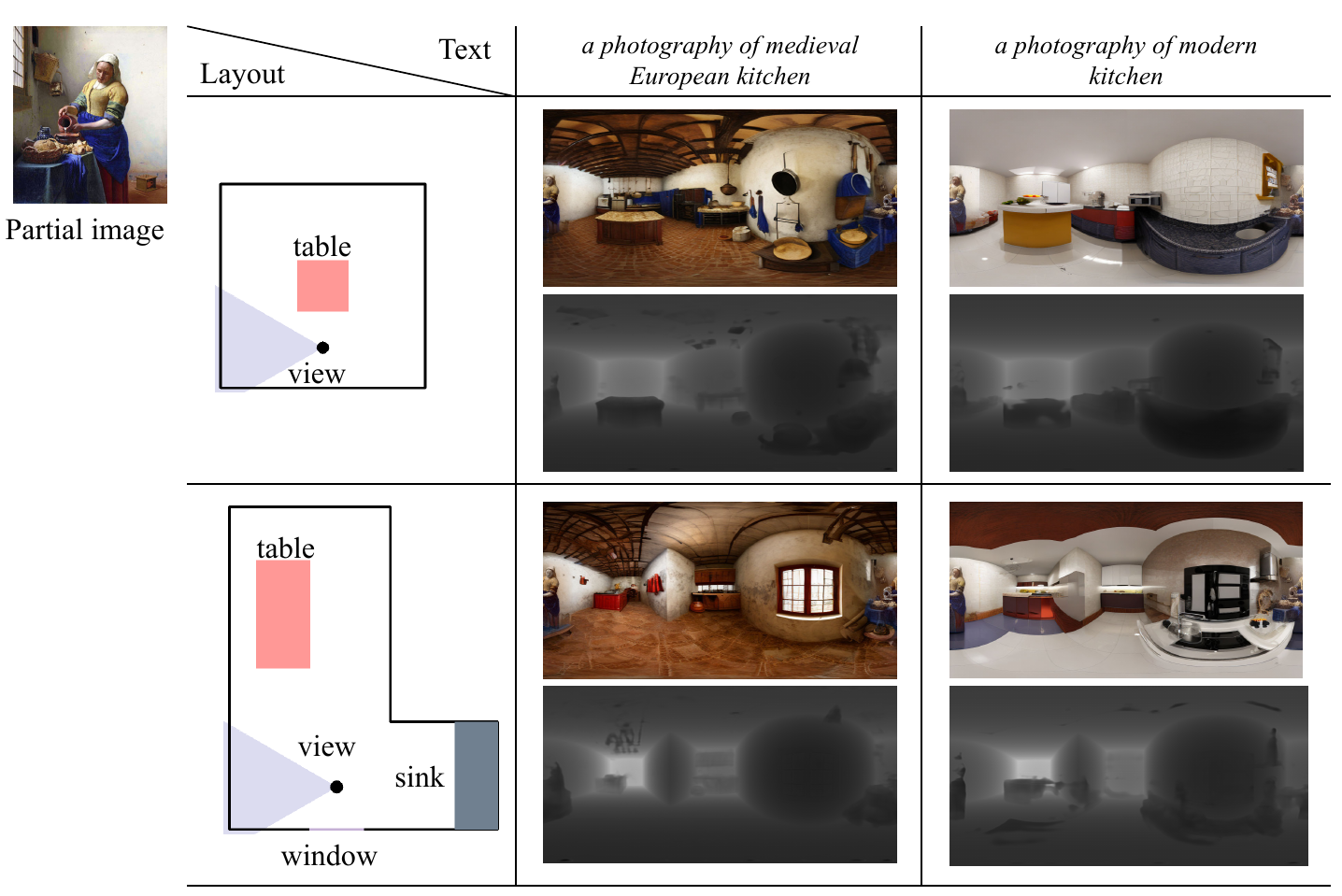}{From a given partial image, layout, and text prompt, our method generates the 360-degree RGB space and depth. We used a painting titled "The Milkmaid" by Johannes Vermeer as a partial image. Various 3D scenes can be generated for the same partial image using different layouts and text prompts.}{150mm}

\putFigWWpdf{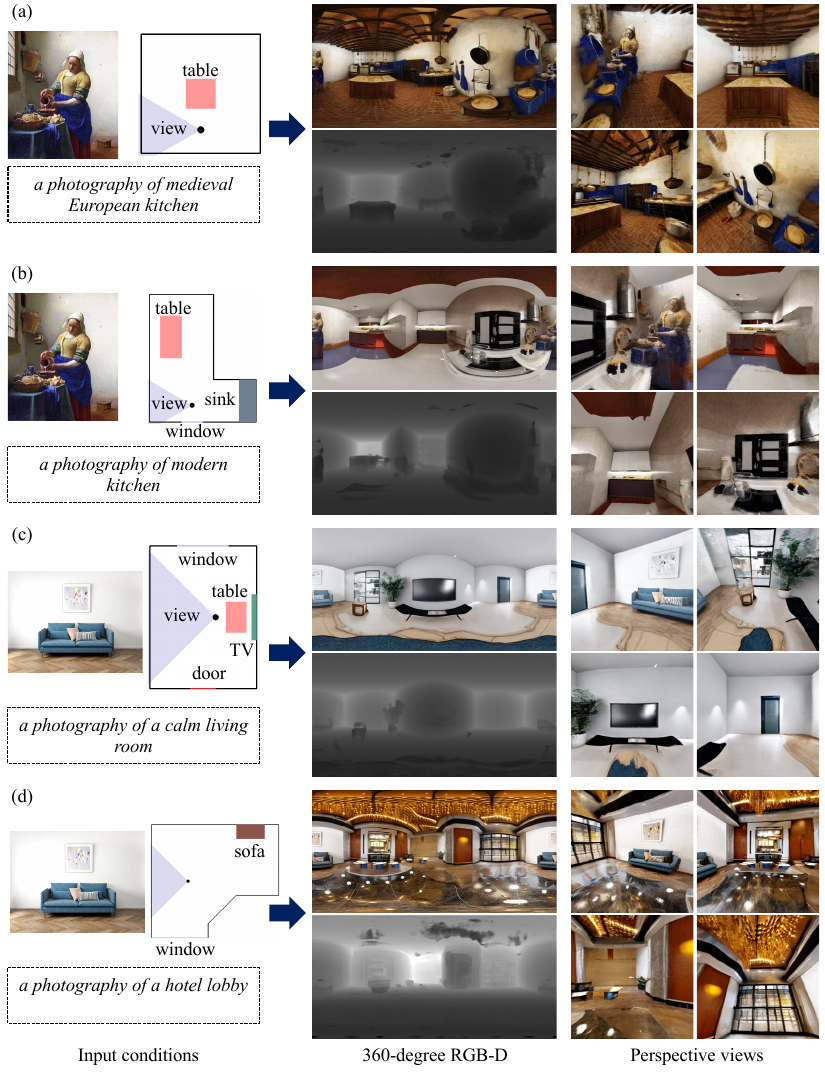}{The various generated indoor 3D scenes represented by 360-degree RGB-D images and free perspective images rendered using NeRF owing to conditions outside the used dataset. (a) (b) We used a painting titled "The Milkmaid" by Johannes Vermeer as a partial image.  (c) (d) A photo of sofas downloaded from the web (https://www.photo-ac.com/) was provided as a partial image.}{160mm}

\putFigWWpdf{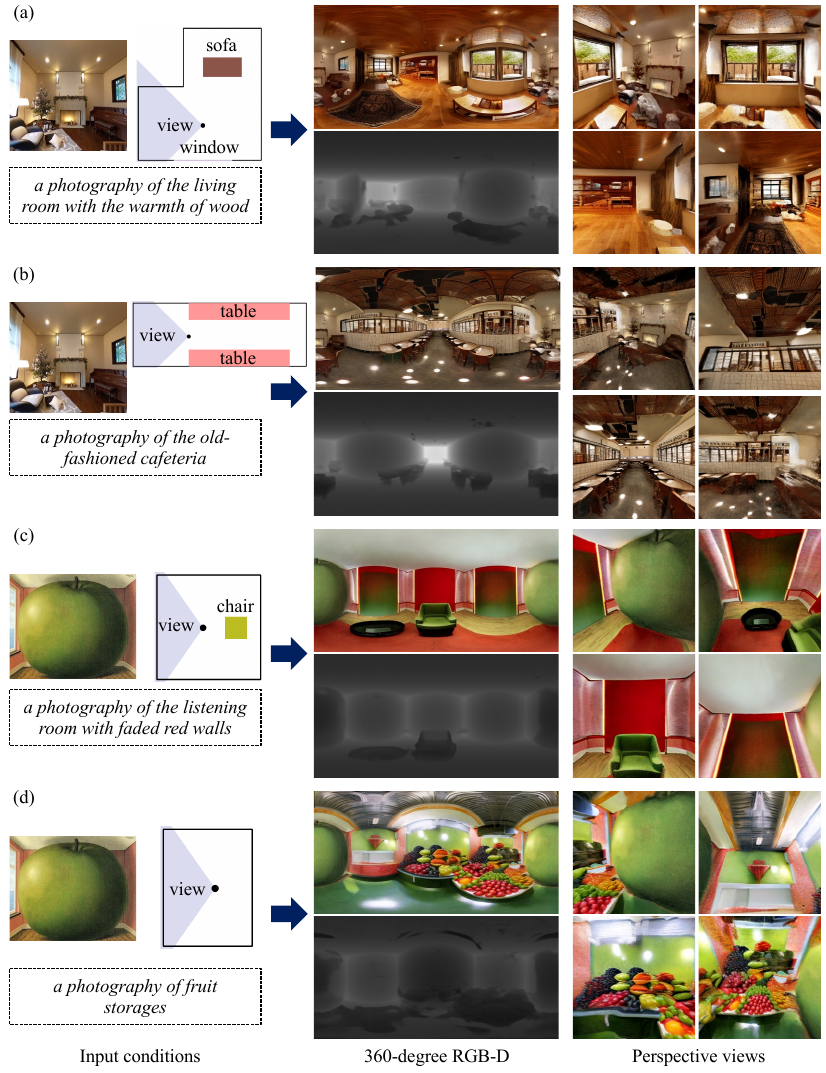}{The various generated indoor 3D scenes represented by 360-degree RGB-D images and free perspective images rendered using NeRF owing to conditions outside the used dataset. (a) (b) An image captured by the author using a camera is shown as a partial image. (e) (f) We presented a painting titled "The Listening Room" by René Magritte as a partial image.}{160mm}

\putFigWWpdf{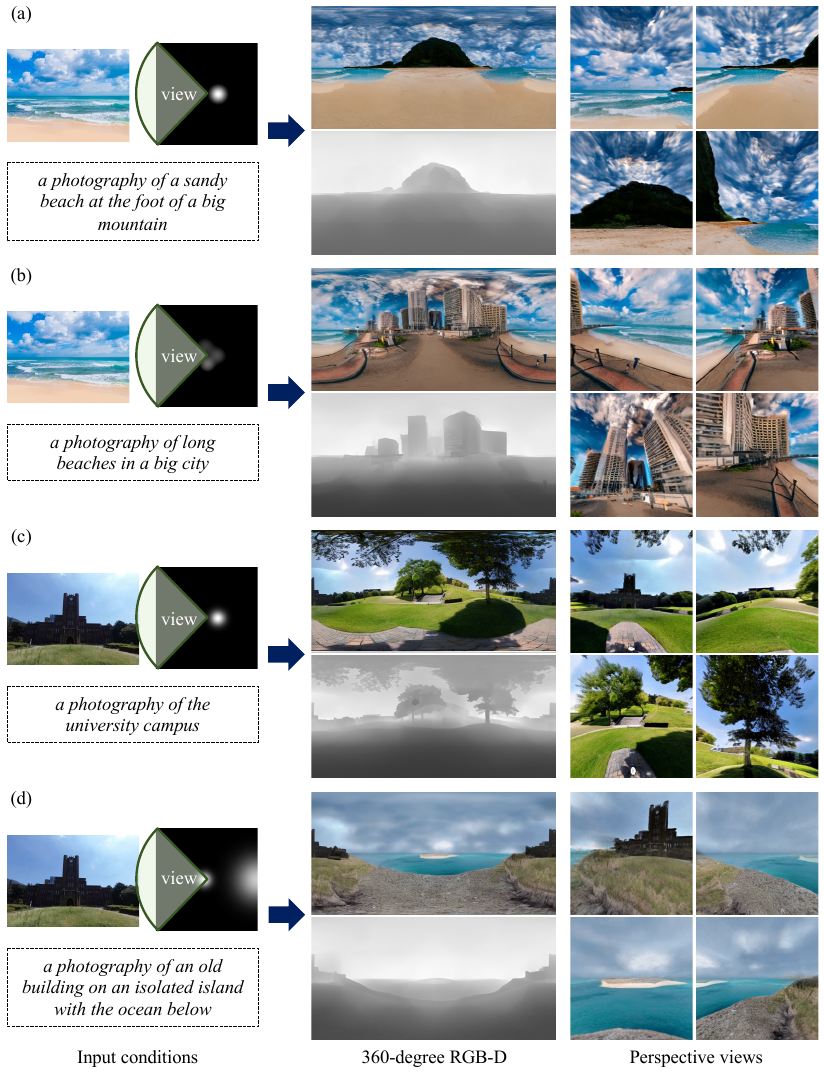}{The various generated outdoor 3D scenes represented by 360-degree RGB-D images and free perspective images rendered using NeRF owing to conditions outside the used dataset. (a) (b) A photo of a sandy beach downloaded from the web (https://www.photo-ac.com/) was given as a partial image. (c) (d) An image captured by the author using a camera is shown as a partial image.}{160mm}

\putFigWWpdf{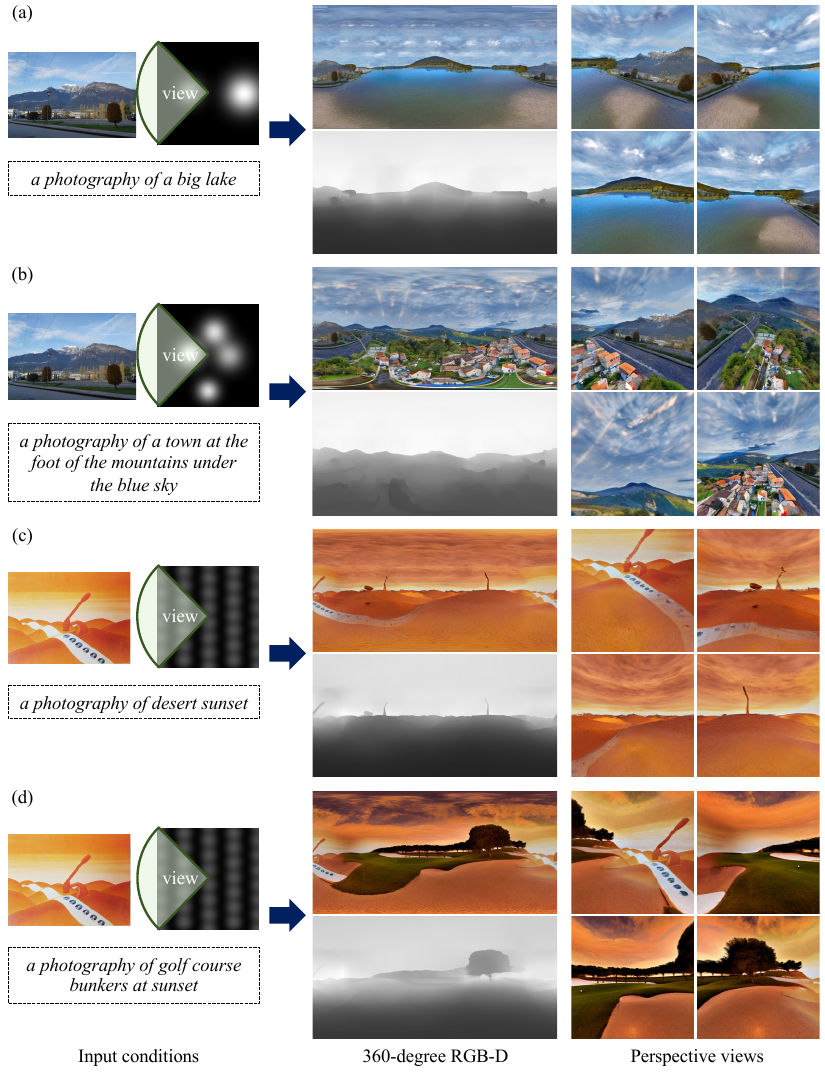}{The various generated outdoor 3D scenes represented by 360-degree RGB-D images and free perspective images rendered using NeRF owing to conditions outside the used dataset. (a) (b) An image captured by the author using a camera is shown as a partial image. (c) and (d) We provided a painting titled "Day after Day" by Jean-Michel Folon as a partial image.}{160mm}

We evaluated the results of the 3D scene generation based on user-generated conditions outside the dataset used for fine-tuning. In this experiment, the end-to-end method was used to estimate the depth in indoor scenes, whereas the depth integration method was applied to outdoor scenes because the SceneDreamer dataset is limited to natural scenery, such as mountainous areas and seashores, using monocular depth estimation models trained on an external dataset. Because CD is effective for fine-tuning with additional text annotations, we used a simpler method without CD in the in-the-wild experiments described in this section. The terrain map $T \in \mathbb{R}^{ H_{\rm{ter}} \times W_{\rm{ter}}}$was created as a mixed Gaussian distribution in the following equation:
\begin{equation}
T_p = \sum_{k=1}^K \pi_k \exp \left( -\frac{1}{2} ( p - \mu_k )^{\top} \Sigma_k^{-1} ( p - \mu_k ) \right),
\end{equation}
where $p \in \{ 1, 2, \cdots, H_{\rm{ter}} \} \times \{ 1, 2, \cdots, W_{\rm{ter}} \} $ is the location on the 2-D map, $K$ is the number of mixtures, and $\pi_k \in \mathbb{R}$, $\mu_k \in \mathbb{R}^2$, and $\Sigma_k \in \mathbb{R}^{2 \times 2}$ are the parameters of the weights, mean, and covariance matrix of the element distribution, respectively.

Additional examples of 3D scenes generated using the proposed method conditioned on text, partial images, and layouts are presented in \cref{fig:sample_variout_cond,fig:sample_free_in1,fig:sample_free_in2,fig:sample_free_out1,fig:sample_free_out2}. In these figures, the aspect ratios of the ERP images were converted to 2:1 for display purposes. These conditions were created freely by the authors. It can be seen that the generated scene contains the given partial image and conforms to the instructions of the text prompt according to the given layout. In addition to the coarse depth created by the room shape or terrain alone, the geometry of objects such as chairs, tables, trees, and buildings can be seen. \cref{fig:sample_variout_cond} shows how various scenes can be generated in a controlled manner by changing the combination of layout and text for the same partial image. \cref{fig:sample_free_in1,fig:sample_free_in2,fig:sample_free_out1,fig:sample_free_out2} shows that our method can generate a variety of 3D scenes from photos on the web, photos taken in the real world, and fanciful paintings, taking into account the layout and text requirements we give. These results show that the proposed method can generate 360-degree RGB-D images with appearance, geometry, and overall context controlled according to the input information, even outside the dataset used for fine-tuning.

\section{Discussion}
\label{sec:discussion}

\subsection{Advantages of Using 360-Degree Images}

\putFigWWpdf{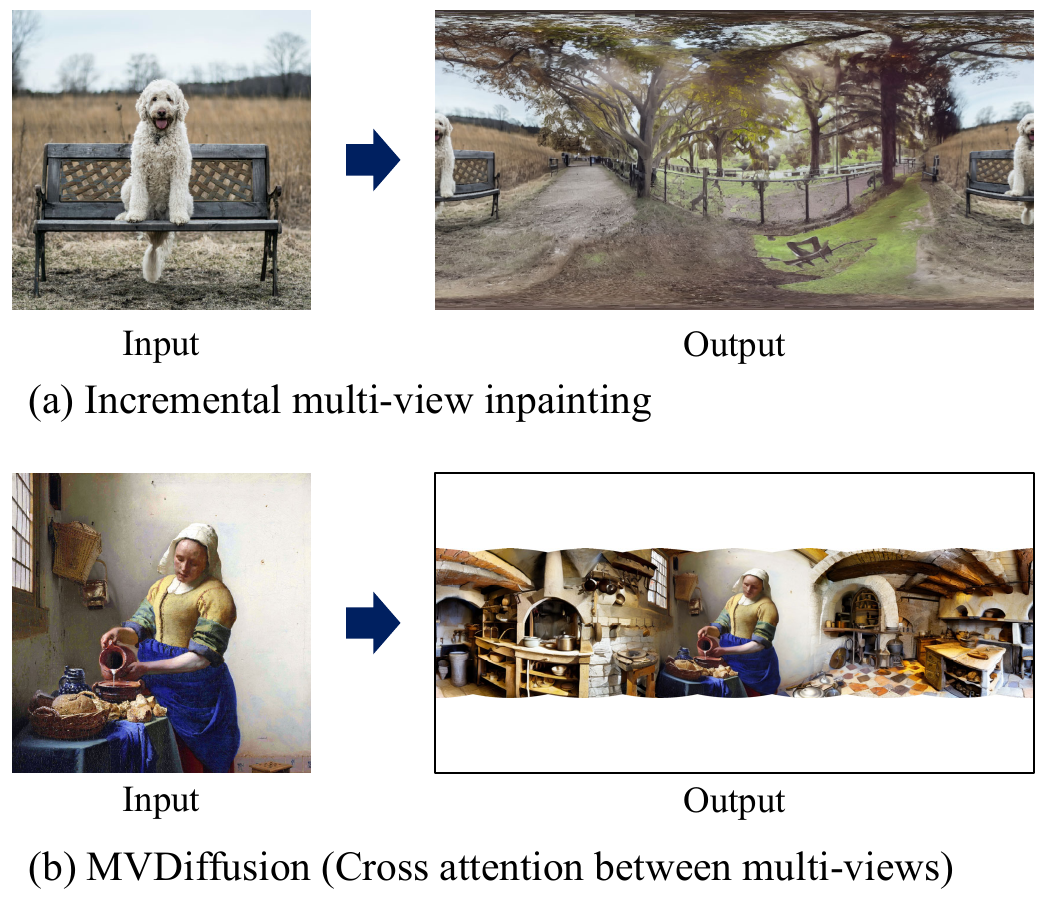}{Examples of the scene generation from a partial image through the generation of perspective projection images. The generated scenes were displayed in ERP format. (a) In incremental multiview inpainting of the perspective image downloaded from the web (https://unsplash.com/@overture\_creations/), the road disappears on the other side, indicating that the scene is not consistent. (b) MVDiffusion maintains consistency between multiple views; however, the computational cost is high because cross attention is required for each combination of multiple views.}{84mm}

The proposed method uses a trained text-to-image model to generate a 2D image, from which the depth is generated. The proposed method is unique because it uses a 360-degree image as the 2D image for generation. Using 360-degree images is advantageous over perspective projection images in terms of scene consistency and reduced computational costs. \cref{fig:advantage_360} shows examples of the generated scene from a partial image by the incremental multi-view inpanting and MVDiffusion \cite{Tang2023mvdiffusion}. Incremental multi-view inpainting is a method of repeating SD inpainting by projecting an input image from a different viewpoint. In the example shown in this figure, the road disappears, indicating that the scene is inconsistent. This is due to the fact that inpainting is performed on each perspective projection image; therefore, the overall consistency cannot be guaranteed. In addition, inpainting must be applied repeatedly, which is computationally expensive and difficult to parallelize. MVDiffusion, on the other hand, takes cross-attention among multiple views and generates multiple views that are simultaneously consistent using SD. This method is computationally expensive because it requires running SD for each view and paying cross-attention to the combinations of multiple views. The order of computational complexity is $O(N^2)$, where $N$ is the number of viewpoints. Because the proposed method generates a single 360-degree image, it is easy to achieve scene consistency at a low computational cost. However, the resolution of the generated image using ERP is lower than that of multiview images, and a higher resolution is a future challenge.

\subsection{Limitation}

\putFigWWpdf{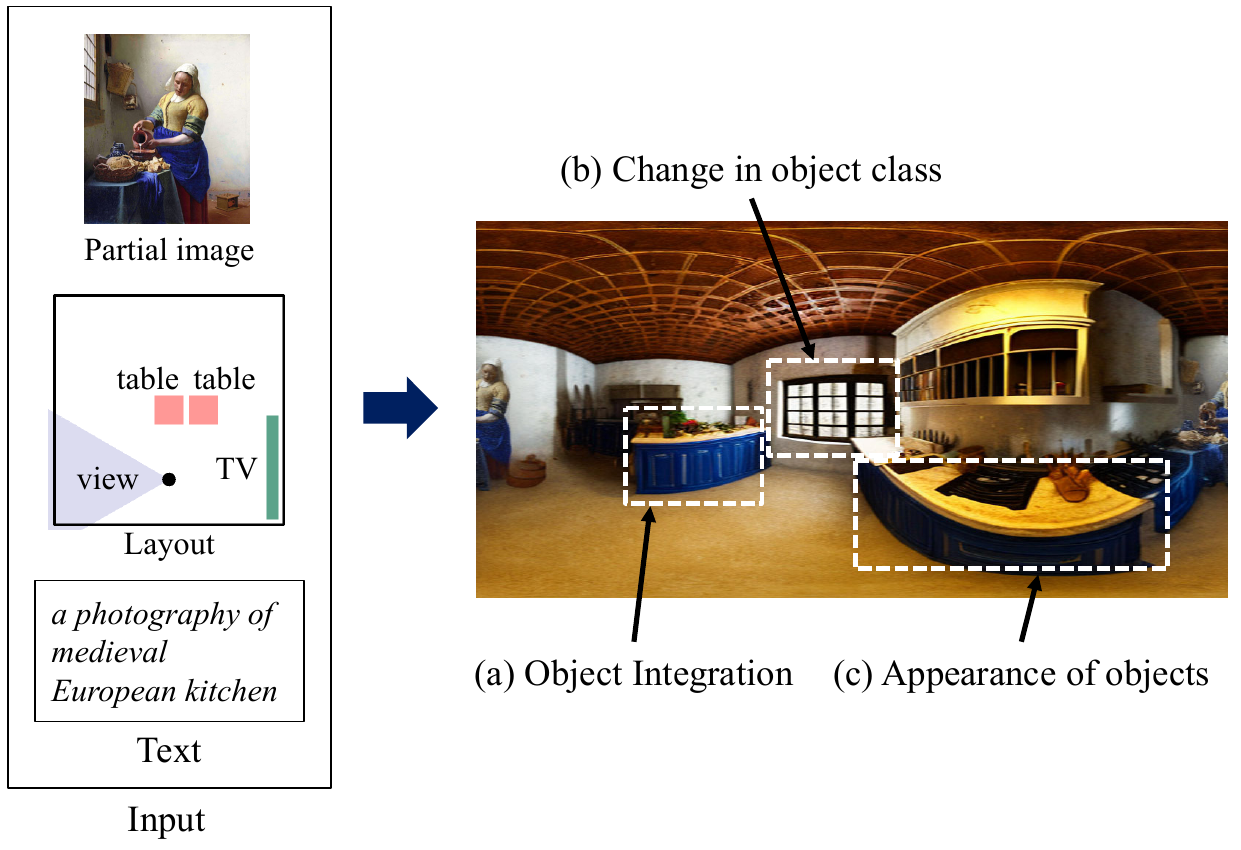}{Examples of limitations of 360-degree RGB-D generation from multimodal conditions. (a) When two tables specified in the layout condition overlap in the ERP, they are merged and generated as a single table. (b) Although the layout conditions dictate the placement of a television, it is generated and converted to a window because it does not conform to the context of ``a medieval European kitchen,'' which is presented in the text prompt. (c) Where nothing is specified in the layout conditions, objects may be generated automatically according to text prompts. It is impossible to specify areas where no objects exist.}{120mm}

\putFigWWpdf{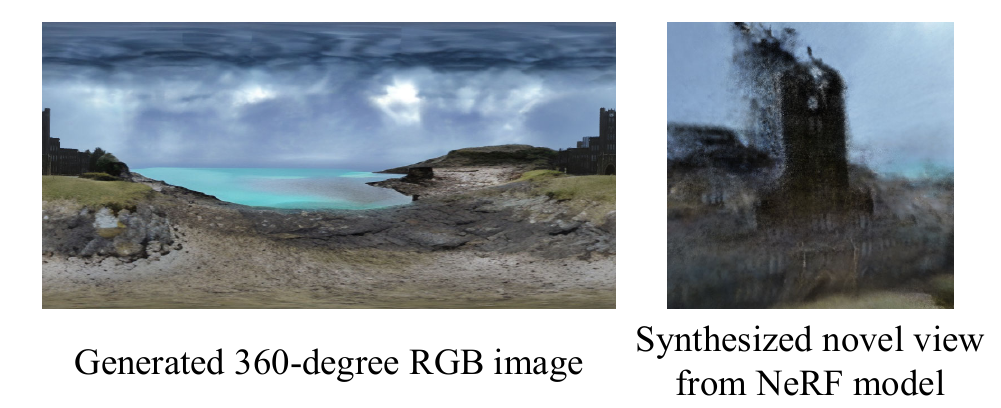}{Examples of limitations of synthesized novel views from the NeRF model trained on the generated 360-degree image. It is difficult to synthesize plausible views when generating 3D scenes from 360-degree RGB-D images with large missing regions that exceed image completion capabilities. In this example, the image quality is significantly reduced in the occluded region at the back of the building.}{96mm}

Although the performance of the proposed method was promising, it had several limitations.

\cref{fig:limitation_360rgbd} shows examples of problems in RGB generation. First, if the objects specified in the layout are in overlapping positions from a viewpoint, they cannot be separated and drawn in the correct number and position. This is because the 2D layout information is converted to ERP for input, which requires additional ingenuity, such as generating a 3D scene jointly from multiple viewpoints. Second, when using conditions outside the dataset, the specified conditions may not be reflected, depending on the interaction between each condition. For example, there is the phenomenon that certain text prompts do not produce certain objects. Third, it is not possible to specify the regions where objects do not exist. Except for the regions where objects are specified, object generation is controlled by other conditions such as partial image, depth, and text. 

\cref{fig:limitation_nerf} shows examples of problems in 6 DoF 3D scene generation. It is difficult to synthesize plausible views when generating 3D scenes from 360-degree RGB-D images with large missing regions that exceed image completion capabilities.

We hope that these limitations will be addressed in future studies.

\end{document}